\documentclass[10pt]{article} % For LaTeX2e
\usepackage[accepted]{tmlr}
\usepackage{hyperref}
\usepackage{url}

\usepackage{algorithm, algpseudocode}
\usepackage{microtype}
\usepackage{graphicx}
\usepackage{subcaption}
\usepackage{booktabs} % for professional tables

\usepackage{amsmath}
\usepackage{amssymb}
\usepackage{mathtools}
\usepackage{amsthm}
\usepackage{macros}
\usepackage{bbm}
\usepackage{wrapfig}
\usepackage{tabularx}
\usepackage{multirow}

\newcolumntype{Y}{>{\centering\arraybackslash}X}

\title{Unsupervised Learning of Neurosymbolic Encoders}

\author{%
\name \hspace{-0.4in}Eric Zhan* \email ezhan@caltech.edu \\
      \addr California Institute of Technology
      \AND
      \name \hspace{-0.4in}Jennifer J. Sun* \email jjsun@caltech.edu \\
      \addr California Institute of Technology
      \AND
      \name \hspace{-0.4in}Ann Kennedy \\
      \addr Northwestern University Feinberg School of Medicine
      \AND
      \name \hspace{-0.4in}Yisong Yue \\
      \addr California Institute of Technology
      \AND
      \name \hspace{-0.4in}Swarat Chaudhuri \\
      \addr University of Texas at Austin 
      \AND
      \hspace{-0.2in}* Equal contribution}

\def\month{11}  % Insert correct month for camera-ready version
\def\year{2022} % Insert correct year for camera-ready version
\def\openreview{\url{https://openreview.net/forum?id=eWvBEMTlRq}} % Insert correct link to OpenReview for camera-ready version

\begin{document}

\maketitle

\begin{abstract}
We present a framework for the unsupervised learning of neurosymbolic encoders, which are encoders obtained by composing neural networks with symbolic programs from a domain-specific language. 
Our framework naturally incorporates symbolic expert knowledge into the learning process, which leads to more interpretable and factorized latent representations compared to fully neural encoders.
We integrate modern program synthesis techniques with the variational autoencoding (VAE) framework, in order to learn a neurosymbolic encoder in conjunction with a standard decoder.
The programmatic descriptions from our encoders can benefit many analysis workflows, such as in behavior modeling where interpreting agent actions and movements is important.
We evaluate our method on learning latent representations for real-world trajectory data from animal biology and sports analytics. 
We show that our approach offers significantly better separation of meaningful categories than standard VAEs and leads to practical gains on downstream analysis tasks, such as for behavior classification. 
Code can be found at \url{https://github.com/ezhan94/neurosymbolic-encoders}.
\end{abstract}

\section{Introduction}
Advances in unsupervised learning have enabled the discovery of latent structures in data from a variety of domains, such as image data~\citep{dupont2018learning}, sound recordings~\citep{calhoun2019unsupervised}, and tracking data~\citep{luxem2020identifying}. 
For instance, a common approach is to use encoder-decoder frameworks, such as variational autoencoders (VAEs)~\citep{kingma2013auto}, to identify a low-dimensional latent representation from the raw data that could contain disentangled factors of variation~\citep{dupont2018learning} or semantically meaningful clusters~\citep{luxem2020identifying}. 
Such approaches typically employ complex mappings based on neural networks, and explaining how the model assigns inputs to latent representations can be challenging \citep{zhang2020survey}.

In this paper, we introduce \emph{unsupervised neurosymbolic representation learning}, which allows part of a representation to be computed using symbolic \emph{encoder programs} written in a predefined domain-specific language (DSL). (The rest of the representation is computed using a neural network.) 
The use of such neurosymbolic encoders can offer two key benefits over purely neural approach. 
First, since a DSL reflects structured domain knowledge, neurosymbolic encoders can often produce representations that are human-interpretable \citep{verma2018programmatically,shah2020learning}. 
Second, as observed in studies that used hand-crafted programmatic encoders \citep{zhan2020learning}, these representations can potentially be more factorized or well-separated into meaningful categories than purely neural representations. 

Our learning algorithm is grounded in the VAE framework \citep{kingma2013auto,mnih2014neural} and aims to discover a neurosymbolic encoder coupled with 
a standard neural decoder.\footnote{Some prior work have studied the complementary problem of learning (neuro-)symbolic decoders (e.g,.~\cite{ellis2017learning,feinman2020learning}).  See Section \ref{sec:related} for more discussion.}
A key challenge here is that the space of programs in a DSL is combinatorial. We tackle this problem by assuming programs to be differentiable and by tightly integrating standard VAE training with modern program synthesis methods \citep{chaudhuri2021neurosymbolic,shah2020learning}. 
We further show how to incorporate ideas from adversarial information factorization \citep{creswell2017adversarial} and enforcing capacity constraints \citep{burgess2018understanding, dupont2018learning} in order to mitigate issues such as posterior and index collapse in the learned representation.

Programmatic descriptions from neurosymbolic encoders are especially useful in behavior analysis \citep{segalin2020mouse,sun2020task}, where domain experts routinely interpret clusters of behaviors as part of an analysis workflow. Accordingly, our experimental evaluation focuses on this setting. 
By integrating domain knowledge using program synthesis, we demonstrate that our clusters are inherently interpretable and better aligned with human-annotated labels across multiple behavior analysis datasets. To validate the end-to-end practicality for analysis workflows, we integrate our automatically learned programs into a state-of-the-art behavior analysis framework, Task Programming \citep{sun2020task}, that typically relies on   expert-crafted programs, and demonstrate competitive performance using our automatically synthesized programs.

To summarize, our contributions are:
\begin{itemize}
    \item We propose a neurosymbolic approach to representation learning, in which part of the latent representation is produced by an interpretable encoder program, while the rest is computed using a  neural network.
    \item We realize the approach via a  learning algorithm that combines VAE training and program synthesis.
    \item We show that our approach can significantly outperform purely neural encoders in extracting semantically meaningful representations of behavior, as measured by standard unsupervised metrics.
    \item We further explore the flexibility of our approach, by showing that performance can be robust across different DSL designs by domain experts. 
    \item We showcase the practicality of our approach on downstream tasks, by incorporating our approach into a state-of-the-art self-supervised learning approach for behavior analysis \citep{sun2020task}.
\end{itemize}

\section{Background}

%%%
\subsection{Variational Autoencoders}\label{sec:vae}

We build on VAEs \citep{kingma2013auto,mnih2014neural}, a latent variable modeling framework shown to learn effective latent representations  (also called encodings/embeddings) \citep{higgins2016beta,zhao2017learning,yingzhen2018disentangled} and can capture the generative process \citep{oord2017neural,vahdat2020nvae,zhan2020learning}.
VAEs introduce a latent variable $\bfz$, an encoder $q_{\phi}$, a decoder $p_{\theta}$, and a prior distribution $p$ on $\bfz$. $\phi$ and $\theta$ are the parameters of the $q$ and $p$ respectively, often instantiated with neural networks. The learning objective is to maximize the evidence lower bound (ELBO) of the data log-likelihood:
\begin{equation}
\begin{aligned}
\textsc{ELBO}:=\ \ \ \Expect_{q_{\phi}(\bfz|\bfx)} \big[ \log p_{\theta}(\bfx | \bfz) \big] - & D_{KL} \bigbr{q_{\phi}(\bfz|\bfx) || p(\bfz)} \leq \quad \log p(\bfx).
\label{eq:vae_obj}
\end{aligned}
\end{equation}
The first term in \eqref{eq:vae_obj} is the log-density assigned to the data, while the second term is the KL-divergence between the prior and approximate posterior of $\bfz$. Latent representations $\bfz$ are often continuous and modeled with a Gaussian prior, but $\bfz$ can be modeled to contain discrete dimensions as well \citep{kingma2014semi, hu2017toward, dupont2018learning}.  Our experiments are focused on behavioral tracking data in the form of trajectories, and so in practice we utilize a trajectory variant of VAEs \citep{co2018self, zhan2020learning, sun2020task}, described in Section \ref{sec:traj}.

One challenge with VAEs (and deep encoder-decoder models in general) is that while the model is expressive, it is often difficult to interpret what is encoded in the latent representation $\bfz$. Common approaches include taking traversals in the latent space and visualizing the resulting generations \citep{burgess2018understanding}, or post-processing the latent variables using techniques such as clustering \citep{luxem2020identifying}.
Such techniques are post-hoc and thus cannot guide (in an interpretable way) the encoder to be biased towards a family of structures.  Some recent work have studied how to impose structure in the form of graphical models or dynamics in the latent space \citep{johnson2016composing,deng2017factorized}, and our work can be thought of as a first step towards imposing structure in the form of symbolic knowledge encoded in a domain specific programming language.

%%%
\subsection{Synthesis of Differentiable Programs}\label{sec:diff_prog_synth}

Our approach utilizes recent work on the synthesis of differentiable programs~\citep{chaudhuri2021neurosymbolic,shah2020learning,valkov2018houdini},  
where one learns both the discrete structure of the  symbolic program (analogous to the architecture of a neural network) as well as the differentiable parameters within that structure.
Our formulation closely follows that of \cite{shah2020learning}. We use a domain-specific programming language (DSL), generated with a context-free grammar (see Figure \ref{fig:dsl} for an example). A program is represented as a pair $(\alpha, \psi)$, where $\alpha$ is a discrete program architecture and $\psi$ are its real-valued parameters. We denote $\calP$ as the space of symbolic programs (i.e. programs with complete architectures). The semantics of a program $(\alpha, \psi)$ is given by a function $\Sem{\alpha}(x, \psi)$ that is guaranteed to be differentiable in both $x$ and $\psi$.

Like \citet{shah2020learning}, we pose the problem of learning differentiable programs as search through a directed program graph $\calG$. The graph $\calG$ models the top-down construction of program architectures $\alpha$ through the repeated firing of rules of the DSL grammar, starting with an \emph{empty} architecture $\alpha_0$ (represented by the ``start'' nonterminal of the grammar). 
The \emph{leaf nodes} of $\calG$ represent programs with \textit{complete} architectures (no nonterminals). Thus, $\calP$ is the set of programs in the leaf nodes of $\calG$.
The other nodes in $\calG$ contain programs with \textit{partial} architectures (has at least one nonterminal). We interpret a program in a non-leaf node as being neurosymbolic, by viewing its nonterminals as representing neural networks with free parameters.
The root node in $\calG$ is the empty architecture $\alpha_0$, interpreted as a fully neural program. 
An edge $(\alpha, \alpha')$ exists in $\calG$ if one can obtain $\alpha'$ from $\alpha$ by applying a rule in the DSL that replaces a nonterminal in $\alpha$.

Program synthesis in this problem setting equates to searching through $\calG$ to find the optimal complete program architecture, and then learning corresponding parameters $\psi$, i.e., to find the optimal $(\alpha,\psi)$ that minimizes a combination of standard training loss (e.g., classification error) and structural loss (preferring ``simpler'' $\alpha$'s).
\citet{shah2020learning} evaluate multiple strategies for solving this problem and finds \emph{informed search using admissible neural heuristics} to be the most efficient strategy (see appendix). Consequently, we adopt this algorithm for our program synthesis task.

\section{Neurosymbolic Encoders}~\label{sec:neurosymbolic_encoders}
\vspace{-0.3in}

\begin{figure*}[t]
    \centering
        \includegraphics[width=0.99\linewidth]{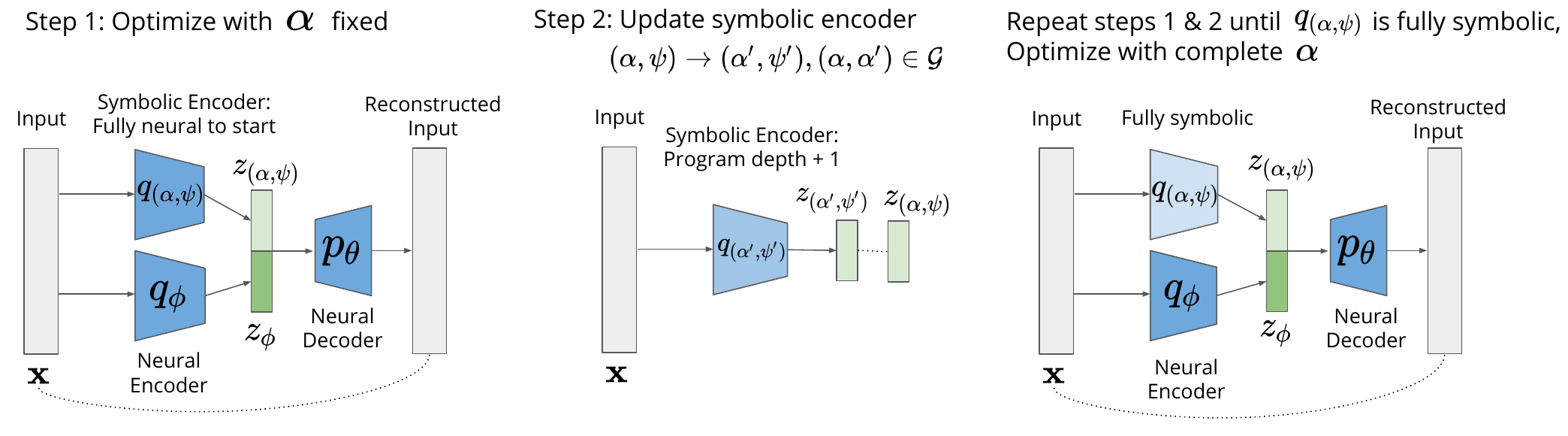}
        \vspace{-0.1in}
        \caption{ \textbf{Learning Neurosymbolic Encoders: Sketch of Algorithm \ref{alg:general_alg}} (Section~\ref{sec:learning_alg}). The symbolic encoder is initially fully neural. We alternate between VAE training with the program architecture fixed (Step 1 as in \eqref{eq:prog_tvae_obj}), and supervised program learning to increase the depth of the program by 1 (Step 2 as in \eqref{eq:update_prog}). Once we reach a symbolic program, we train the model one last time to learn all the parameters. The color (in terms of lightness) of the symbolic encoder corresponds to the encoder becoming more symbolic over time.}
    \label{fig:model}
\end{figure*}

The structure of our neurosymbolic encoder is shown in the right diagram of Figure \ref{fig:model}.
The latent representation $\bfz = [\Zneural, \Zsymb]$ is partitioned into neurally encoded $\Zneural$ and programmatically encoded $\Zsymb$.  This approach boasts several advantages:
\vspace{-0.05in}
\begin{itemize}
    \item The symbolic component of the latent representation is programmatically interpretable.
    \item The neural component can encode any residual information not captured by the program, which maintains the model's capacity compared to deep encoders (see synthetic experiment in Section \ref{sec:q1}).
    \item By incorporating a modular design, we can leverage state-of-the-art learning algorithms for both differentiable encoder-decoder training and program synthesis.
\end{itemize}

We denote $\Qneural$ and $\Qsymb$ as the neural and symbolic encoders respectively (see Figure \ref{fig:model}), where $\Zneural \sim \Qneural(\cdot | \bfx)$ and $\Zsymb \sim \Qsymb(\cdot | \bfx)$. $\Qneural$ is instantiated with a neural network, but $\Qsymb$ is a differentiable program with architecture $\alpha$ and parameters $\psi$ in some program space $\calP$ defined by a DSL. Given an unlabeled training set of $\x$'s, our neurosymbolic-VAE (ns-vae) learning objective becomes:
\begin{equation}
\begin{aligned}
\quad \max_{\PHIneural, (\alpha, \psi), \theta} \quad & \calL_{\text{ns-vae}} (\PHIneural, \alpha, \psi, \theta) \\
= \max_{\PHIneural, (\alpha, \psi), \theta} \quad & \BigExpect_{q_{\PHIneural}(\Zneural|\bfx)q_{(\alpha, \psi)}(\Zsymb|\bfx)} \big[ \underbrace{\log p_{\theta}(\bfx | \Zneural, \Zsymb)}_{\text{reconstruction loss}} \big] \\
 & \quad - \underbrace{D_{KL} \bigbr{\Qneural(\Zneural |\bfx) || p(\Zneural)}}_{\text{regularization for neural latent}} - \underbrace{D_{KL} \bigbr{q_{(\alpha, \psi)}(\Zsymb|\bfx) || p(\Zsymb)}}_{\text{regularization for symbolic latent}}.
\end{aligned}
\label{eq:prog_tvae_obj}
\end{equation}
Compared to the standard VAE objective in \eqref{eq:vae_obj} for a single neural encoder, \eqref{eq:prog_tvae_obj} has separate KL-divergence terms for the neural and  programmatic encoders. 

\subsection{Learning Algorithm}~\label{sec:learning_alg}
\vspace{-0.2in}

The challenge with solving for \eqref{eq:prog_tvae_obj} is that while $(\PHIneural, \psi, \theta)$ can be optimized via back-propagation with $\alpha$ fixed, optimizing for $\alpha$ is a discrete optimization problem.
Since it is difficult to jointly optimize over both continuous and discrete spaces, we take an iterative, alternating optimization approach.  We start with a fully neural program (one with empty architecture $\alpha_0$ as described in Section \ref{sec:diff_prog_synth}) trained using standard differentiable optimization (Figure \ref{fig:model}, Step 1). We then gradually make it more symbolic (Figure \ref{fig:model}, Step 2) by finding a program that is a child of the current program in $\calG$ (more symbolic by construction of $\calG$) that outputs as similar to the current latent representations as possible:
\begin{equation}
\begin{aligned}
\min_{\alpha': (\alpha, \alpha') \in \calG,\ \psi'} \quad & \calL_{\text{supervised}} \bigbr{q_{(\alpha, \psi)}(\bfx), q_{(\alpha', \psi')}(\bfx)},
\end{aligned}
\label{eq:update_prog}
\end{equation}
which can be viewed as a form of distillation (from less symbolic to more symbolic programs) via matching the input/output behavior.  
We solve \eqref{eq:update_prog} by enumerating over all child programs of the current search tree and selecting the best one, which is similar to one iteration of iteratively-deepened depth-first search in \citet{shah2020learning} (more details in Section \ref{sec:near_details}). We alternate between optimizing \eqref{eq:prog_tvae_obj} and \eqref{eq:update_prog} until we obtain a complete program. Algorithm \ref{alg:general_alg} outlines this procedure and is guaranteed to terminate if $\calG$ is finite by specifying a maximum program depth.

We chose this optimization procedure for two reasons.  First, it maximally leverages  state-of-the-art tools in both differentiable latent variable modeling (VAE-style training) and supervised program synthesis (for distillation), leading to tractable algorithm design.  Second, this procedure never makes a drastic change to the program architecture, leading to relatively stable learning behavior across iterations.

\begin{minipage}{0.48\textwidth}
    \begin{algorithm}[H]
    \caption{Learning a neurosymbolic encoder}
    \label{alg:general_alg}
    \begin{small}
    \begin{algorithmic}[1]
        \State \textbf{Input}: program space $\calP$, program graph $\calG$
        \State initialize $\PHIneural, \psi, \theta, \alpha = \alpha_0$ (empty architecture)
        \While{$\alpha$ is not complete}
          \State $\PHIneural, \psi, \theta \leftarrow$ optimize \eqref{eq:prog_tvae_obj} with $\alpha$ fixed
          \State $(\alpha, \psi) \leftarrow$ optimize \eqref{eq:update_prog}
        \EndWhile
        \State $\PHIneural, \psi, \theta \leftarrow$ optimize \eqref{eq:prog_tvae_obj} with complete $\alpha$
        \State \textbf{Return}: encoder $\{ \Qneural, q_{(\alpha, \psi)} \}$
    \end{algorithmic}
    \end{small}
    \end{algorithm}
\end{minipage}
\begin{minipage}{0.48\textwidth}
    \begin{algorithm}[H]
    \caption{Learning a neurosymbolic encoder \\ with $k$ programs}
    \label{alg:mutliple_alg}
    \begin{small}
    \begin{algorithmic}[1]
        \State \textbf{Input}: program space $\calP$, program graph $\calG$, $k$
        \For{$i = 1..k$}
          \State fix programs $\{ q_{(\alpha_1, \psi_1)}, \dots, q_{(\alpha_{i-1}, \psi_{i-1})} \}$
          \State execute Algorithm \ref{alg:general_alg} to learn $q_{(\alpha_i, \psi_i)}$
          \State remove $q_{(\alpha_i, \psi_i)}$ from $\calP$ to avoid redundancies
        \EndFor
        \State \textbf{Return}: encoder $\{ \Qneural, q_{(\alpha_1, \psi_1)}, \dots, q_{(\alpha_k, \psi_k)} \}$
    \end{algorithmic}
    \end{small}
    \end{algorithm}
\end{minipage}

\subsection{Program Synthesis via NEAR}
\label{sec:near_details}

Our strategy for solving \eqref{eq:update_prog} utilizes the setup in \cite{shah2020learning}. We summarize the key points below.

\begin{wrapfigure}{r}{0.38\textwidth}
    \vspace{-10pt}
    \centering
    \includegraphics[width=\linewidth]{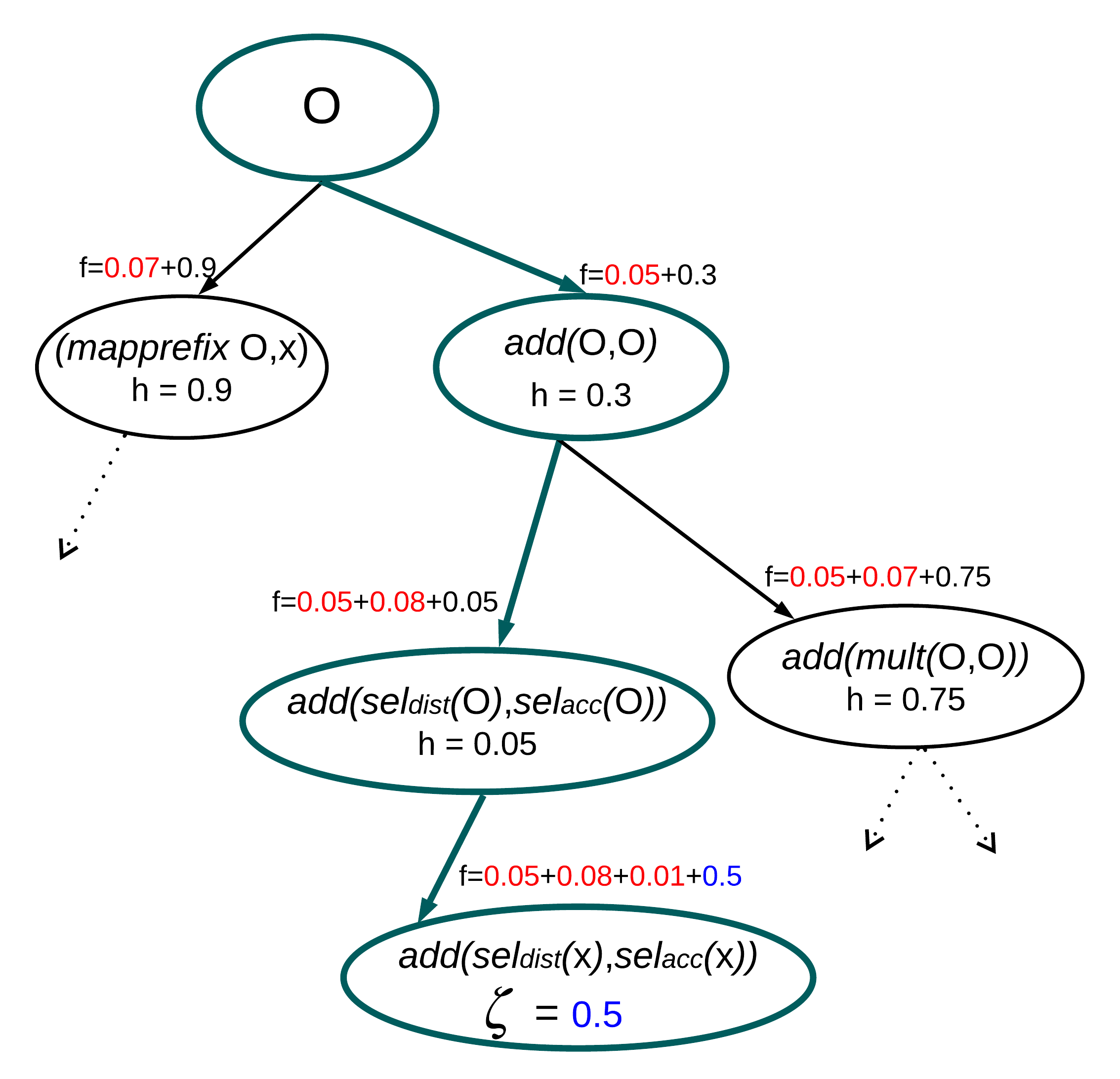}
    \caption{Figure 4 from \cite{shah2020learning}. Structural costs $s$ are in red, heuristic values $h$ in black, and prediction errors $\zeta$ in blue. O refers to nonterminals.}
    \label{fig:near_program_graph}
    \vspace{-50pt}
\end{wrapfigure}

\textbf{Program graph $\calG$.}
\cite{shah2020learning} learns programs in a supervised learning setting that minimizes a structural cost $s$ (deeper programs are more costly) and a prediction error $\zeta$: 
\begin{equation}
(\alpha^*, \psi^*) = \argmin_{(\alpha, \psi)} (s(\alpha) + \zeta(\alpha,\psi)).
\label{eq:near_objective}
\end{equation}
\cite{shah2020learning} construct a program graph $\calG$ such that solving \eqref{eq:near_objective} equates to finding a leaf node with the minimum path cost on $\calG$. We include a copy of their illustration for $\calG$ in Figure \ref{fig:near_program_graph}. Our problem definition in \eqref{eq:update_prog} is very similar so we utilize the same program graph. The difference is that our labels are not ground-truth but rather the labels assigned by the current neurosymbolic encoder.

\textbf{Neural heuristic $h$.}
\cite{shah2020learning} solve \eqref{eq:near_objective} by introducing a heuristic as a neural admissible relaxation (NEAR for short). Leveraging a fully differentiable DSL, they use neural networks to fill in for nonterminals in programs and show that the performance of such neurosymbolic programs are underestimates of the total path costs of descendent leaf nodes and thus, can be used as an admissible heuristic. This allows them to integrate their heuristic with several graph search algorithms, of which they adopt A* search and iteratively-deepened depth-first-search with branch-and-bound (IDS-BB). We use IDS-BB in our work.

\textbf{IDS-BB.}
The full algorithm for IDS-BB is described in Algorithm 2 in \cite{shah2020learning}. In our work, this reduces to the following: 
\begin{enumerate}
    \item For the current program, we enumerate its children in $\calG$.
    \item We compute the heuristic for each child in $\calG$ by replacing any nonterminals with neural networks.
    \item We commit to the most promising child (with respect to the heuristic) and update the program, which can be viewed as one iteration of the full IDS-BB algorithm.
\end{enumerate}

One key difference is that the original IDS-BB algorithm maintains a frontier ordered by the best heuristics encountered so far. However, our label distributions can change between iterations (since the symbolic component of the encoder is updated and thus, so are the labels it assigns), which invalidates the heuristics computed from previous iterations. This leaves a very interesting direction for future work.

\subsection{Learning Multiple Programs}
\label{sec:multiple_progs}

The interpretability of latent representations induced by symbolic encoders $\Qsymb$ ultimately depends on the DSL. For instance, a program that encodes to one of ten classes may not be very interpretable if it involves a matrix multiplication within the program. Instead, we learn \textit{binary} programs that encode sequences into one of two classes (using binary cross-entropy for $\calL_{\text{supervised}}$, a uniform prior on 2-dimensional $\bfz_{(\alpha, \psi)}$, and Gumbel-Softmax \citep{jang2017categorical} to sample $\bfz_{(\alpha, \psi)}$ from the posterior). Figures \ref{fig:example_mouse} \& \ref{fig:example_mouse_2} depict learned binary programs that encode mice trajectories and their interpretations.

To encode more than two classes, we simply learn multiple binary programs by extending \eqref{eq:prog_tvae_obj} to sum over $\calL_{\text{supervised}}$ for $k$ symbolic programs $\{ q_{(\alpha_1, \psi_1)}, \dots, q_{(\alpha_k, \psi_k)} \}$ and corresponding latent representations $\{ \bfz_{(\alpha_1, \psi_1)}, \dots, \bfz_{(\alpha_k, \psi_k)} \}$. This results in $2^k$ classes and a solution space that now scales exponentially (e.g. $|\calP|^k$). Algorithm \ref{alg:mutliple_alg} outlines our greedy solution that reuses Algorithm \ref{alg:general_alg} by iteratively learning one symbolic program at a time. We leave the exploration of more sophisticated search methods as future work.

%%%
\subsection{Dealing with Posterior and Index Collapse}

Deep latent variable models, especially those with discrete latent variables, are notoriously prone to both posterior \citep{bowman2015generating, chen2016variational, oord2017neural} and index \citep{kaiser2018fast} collapse. Since our algorithms optimize for such models repeatedly, they can be susceptible to these failure modes. Below, we summarize two strategies that we found to work well in our setting.\footnote{There are many approaches available for tackling both these issues, but we emphasize that these contributions are orthogonal to ours; as techniques for preventing posterior and index collapse improve, so will the robustness of our algorithm.}

\textbf{Adversarial information factorization.}
Index collapse is the phenomenon in which all data is encoded into one class, resulting in a discrete latent variable $\Zsymb$ that is effectively meaningless.
\citet{creswell2017adversarial} counteracts index collapse by introducing an adversarial network $A_{\omega}$ and maximizing the adversarial loss below to ensure that the adversary $A_{\omega}$ cannot successfully predict $\Zsymb$ from $\Zneural$.

\begin{equation}
\begin{aligned}
\quad \max_{\PHIneural, (\alpha, \psi), \theta} \quad & \calL_{\text{fac}} (\PHIneural, \alpha, \psi, \theta) \\
= \max_{\PHIneural, (\alpha, \psi), \theta} \quad & \BigExpect_{q_{\PHIneural}(\Zneural|\bfx)q_{(\alpha, \psi)}(\Zsymb|\bfx)} \big[ \log p_{\theta}(\bfx | \Zneural, \Zsymb) + \underbrace{\min_{\omega}  \calL_{\text{adv}} \big( A_{\omega}(\Zneural),\Zsymb \big)}_{\text{adversary}} \big] \\
 & \quad - D_{KL} \bigbr{\Qneural(\Zneural |\bfx) || p(\Zneural)} - D_{KL} \bigbr{q_{(\alpha, \psi)}(\Zsymb|\bfx) || p(\Zsymb)} 
\end{aligned}
\label{eq:adv_info_fac}
\end{equation}

\textbf{Channel capacity constraint.} 
Posterior collapse is the phenomenon in which the posterior trivially matches the prior exactly (a KL-divergence of 0) but the latent variables are unused by the decoder.
\cite{burgess2018understanding} and \cite{dupont2018learning} instead force the KL-divergence terms to match capacities $C_{\phi}$ and $C_{(\alpha, \psi)}$, which are hyperparameter (see appendix). Since the KL-divergence is an upper bound on the mutual information between latent variables and the data \citep{kim2018disentangling, dupont2018learning}, this encourages the latent variables to encode information and aims to prevent posterior collapse.
\begin{equation}
\begin{small}
\begin{aligned}
\quad \max_{\PHIneural, (\alpha, \psi), \theta} \quad & \calL_{\text{cap}} (\PHIneural, \alpha, \psi, \theta) \\
= \max_{\PHIneural, (\alpha, \psi), \theta} \quad & \BigExpect_{q_{\PHIneural}(\Zneural|\bfx)q_{(\alpha, \psi)}(\Zsymb|\bfx)} \big[ \log p_{\theta}(\bfx | \Zneural, \Zsymb) \big] \\
&- \gamma_{\phi} \lvert D_{KL} \bigbr{\Qneural(\Zneural |\bfx) || p(\Zneural)} - C_{\phi} \rvert - \gamma_{(\alpha, \psi)} \lvert D_{KL} \bigbr{q_{(\alpha, \psi)}(\Zsymb|\bfx) || p(\Zsymb)} - C_{(\alpha, \psi)} \rvert
\end{aligned}
\label{eq:channel_capacity}
\end{small}
\end{equation}
In our algorithms, we augment our initial objective in \eqref{eq:prog_tvae_obj} with \eqref{eq:adv_info_fac} and \eqref{eq:channel_capacity}:
\begin{equation}
\begin{small}
\max_{\PHIneural, (\alpha, \psi), \theta} \calL_{\text{ns-vae}} (\PHIneural, \alpha, \psi, \theta) + \lambda_{\text{fac}} \calL_{\text{fac}} (\PHIneural, \alpha, \psi, \theta) + \lambda_{\text{cap}} \calL_{\text{cap}} (\PHIneural, \alpha, \psi, \theta), \\
\label{eq:full_obj}
\end{small}
\end{equation}
where $ \lambda_{\text{fac}} =  \lambda_{\text{cap}} = 1$ in our experiments.

%%%
\subsection{Instantiation for Sequential Domains}
\label{sec:traj}
The objective in \eqref{eq:prog_tvae_obj} describes a general problem that is applicable to any domain. In our experiments, we focus on  sequential trajectory data. 
Trajectory data is often used in scientific applications where interpretability is desirable, such as behavior discovery~\citep{luxem2020identifying,hsu2020b}. The ability to easily explain the learned latent representation using programs can help domain experts better understand the structure in their data. Trajectory data is also often relatively low dimensional, which helps experts encode domain knowledge into the DSL more easily  \citep{shah2020learning,sun2020task,zhan2020learning}.

In this domain, $\bfx$ is a trajectory of length $T$: $\bfx = \{ x_1, \dots, x_T \}$. We then factorize the log-density in \eqref{eq:prog_tvae_obj} as a product of conditional probabilities:
\begin{equation}
\log p_{\theta}(\bfx | \Zneural, \Zsymb) = \sum_{t=1}^{T} \log p_{\theta}(x_t | x_{<t}, \Zneural, \Zsymb).
\label{eq:cond_prob}
\end{equation}
When $\Qneural$ and $p_{\theta}$ are instantiated with recurrent neural networks (RNN), the model is more commonly known as a trajectory-VAE (TVAE) \citep{co2018self}.

As the symbolic encoder $\Qsymb$ maps sequences to vectors, we adopt a DSL (Figure \ref{fig:dsl}) previously used for sequence classification \citep{shah2020learning}. Our DSL is purely functional and contains both basic algebraic operations and parameterized library functions. Domain experts can easily augment the DSL with their own functions, such as selection functions that select subsets of features that they deem potentially important. We ensure that all programs in our DSL are differentiable by utilizing a smooth approximation of the $\ifc$-$\thenc$-$\elsec$ construct~\citep{shah2020learning}. Figures \ref{fig:example_mouse} \& \ref{fig:example_mouse_2} depict example programs  (full details in the  appendix).

\begin{figure}[t]
$$
{
\begin{array}{lll}
\alpha & ::= & x \mid \oplus(\alpha_1, \dots, \alpha_k) \mid \oplus_\theta (\alpha_1, \dots, \alpha_k) \\
& & \ifc~\alpha_1~\thenc~\alpha_2~\elsec~\alpha_3 
\mid \select_S~x \mid \mapavg~(\func~x_1. \alpha_1)~x  
\end{array}
}
$$
\vspace{-10pt}
\caption{ Our DSL for sequential domains, similar to the one used in \citet{shah2020learning}. $x$, $\oplus$, and $\oplus_\theta$ represent inputs, basic algebraic operations, and parameterized library functions, respectively. $\func~x. e(x)$ represents a function that evaluates an expression $e(x)$ over the input $x$. $\select_S$ selects a subset $S$ of the dimensions of the input $x$. $\mapavg~g~x$ applies the function $g$ to every element of the sequence $x$ and returns the average of the results. We employ a differentiable approximation of the $\ifc$-$\thenc$-$\elsec$ construct.}
\label{fig:dsl}
\end{figure}

\section{Experiments}

We take a multi-faceted approach to evaluate our unsupervised learning approach using synthetic data and real-world data from animal behavior and sports analytics. 
We also show the end-to-end practicality of our programs by applying them to a downstream behavior classification framework. Our research questions are:

\begin{itemize}
    \item \textbf{Q1: Are the clusters created with our programs meaningful?} (Section~\ref{sec:q1}). We evaluate this aspect both qualitatively and quantitatively by comparing with the truth generative process on synthetic datasets, as well as by comparing to human annotated labels on real-world datasets.
    \item \textbf{Q2: How sensitive is our approach to different DSL choices?} (Section~\ref{sec:q2}). We compare programs learned in our framework from three different DSLs designed by three domain experts for studying animal behavior. The three DSLs (DSL 1, DSL 2, DSL 3) mainly differ in the behavioral features chosen by experts, and are described in Appendix Section C. 
    \item \textbf{Q3: Are the programs useful for downstream tasks?} (Section~\ref{sec:q3}). Ultimately, the practicality of these methods must be validated by their usefulness in downstream tasks such as those used in scientific analyses.  We apply our unsupervised programs to a behavior classification framework called task programming~\citep{sun2020task}. This framework uses hand-crafted programs for self-supervision, which we replace with our automatically learned programs. 
\end{itemize}

\subsection{Experimental Setup}
\label{sec:exp_setup}

\subsubsection{Datasets}

We summarize the datasets used in our experiments, and provide full details in the appendix.

\textbf{Synthetic.} We generate synthetic trajectories by sampling initial positions and velocities from a Gaussian distribution and introducing 2 ground-truth factors of variation as large external forces in the positive/negative x/y directions that affect velocity, totaling to 4 discrete classes.
Velocities are sampled and fixed for the entire trajectory, but we also sample small Gaussian noise at each timestep.
We generate 10k/2k/2k trajectories of length 25 for train/validation/test. 
Figure \ref{fig:synthetic_samples} shows 50 trajectories from the training set.  This dataset is useful because we can evaluate whether our algorithm can learn programs that match the ground-truth factors of variation (such ground-truth information is not available in real-world datasets).

\textbf{CalMS21.} Our primary real-world dataset is the CalMS21 dataset~\citep{sun2021multi}, containing trajectories of socially interacting mice captured for neuroscience experiments. Each frame contains 7 tracked keypoints for each of two mice. The dataset has one set of unlabeled tracking data, which we use to train our neurosymbolic encoder, and another set annotated with 4 labels at each frame by human experts (frame-level behaviors), which we use to evaluate our programs. These labels consists of three behaviors-of-interest between mice (attack, mount, investigation), and a label corresponding to all other behaviors (other), with a more detailed description in~\cite{sun2021multi}. Specifically, our evaluation uses labels from the test split of the CalMS21 classification task. We have 231k/52k/262k trajectories of length 21 for train/val/test. The features in our DSL are selected by a domain expert based on the attributes from~\citet{segalin2020mouse}.
% from a tracker~\citep{segalin2020mouse}
%One mouse is the ``resident" mouse in its homecage and the other is the ``intruder" mouse introduced into the cage. 

\textbf{Basketball.} We use the same basketball dataset as in \citet{shah2020learning} and \citet{zhan2020learning} that tracks professional basketball players. Each trajectory is of length 25 over 8 seconds and contains the $xy$-positions of 10 players. We split trajectories by grouping offensive and defensive players (5 each), effectively doubling the dataset size. We evaluate our algorithm and the baselines with respect to the labels of offensive/defensive players. Our DSL includes additional domain features like player speed and distance-to-basket. In total, we have 177k/31k/27k trajectories for train/val/test.

%%%
\subsubsection{Quantitative Evaluation Setup}

The quantitative evaluations are used to compare our neurosymbolic encoders with baseline unsupervised learning methods on the real-world datasets.

\textbf{Baselines.} We compare our model containing a neurosymbolic encoder against other approaches based on VAEs. In particular, we compare against JointVAE~\citep{dupont2018learning}, which also has both discrete and continuous latent representations, and can be viewed as a fully neural version of our neurosymbolic encoder. Other baselines include VAE, VAE with K-means loss~\citep{ma2019learning,luxem2020identifying}, and Beta-VAE~\citep{burgess2018understanding}. These models have a fully neural encoder and learn continuous latent representations, which we can then use to produce clusters with K-means clustering~\citep{lloyd1982least}. We additionally compare against VQ-VAEs~\citep{oord2017neural}, which produce discrete latent clusters.
We use the TVAE version of all baselines (details included in the appendix).

\begin{figure*}[t]
    \vspace{-0.2in}
    \centering
    \begin{subfigure}[t]{0.25\linewidth}
        \centering
        \raisebox{-58pt}{\includegraphics[width=\columnwidth]{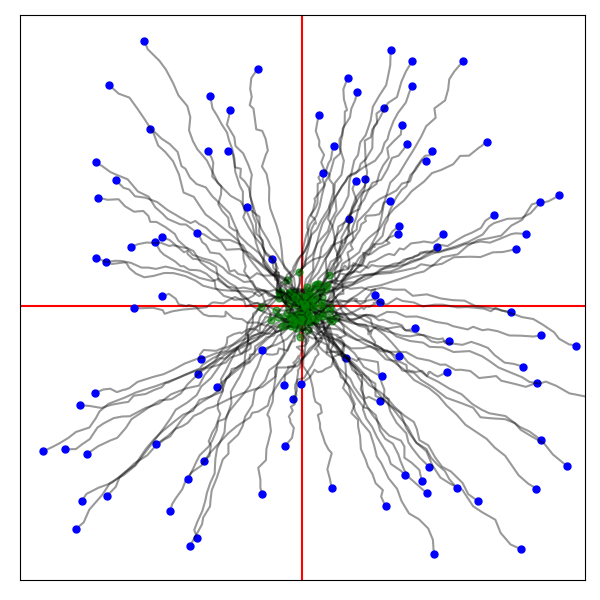}}
        \caption{50 synthetic trajectories}
        \label{fig:synthetic_samples}
    \end{subfigure}
    \begin{subfigure}[t]{0.2\linewidth}
    \centering
        \vspace{-33pt}
        {\small 
        $$
        \begin{array}{l}
        \mathbbm{1}_{[>6.34]} [ \\
        \quad \select_{\mathit{FinalXPosition}}~x ]
        \end{array}
        $$
        }
        \vspace{10pt}
        {\small 
        $$
        \begin{array}{l}
        \mathbbm{1}_{[>8.99]} [ \\
        \quad \select_{\mathit{FinalYPosition}}~x ]
        \end{array}
        $$
        \vspace{8pt}
        }
        \caption{learned programs}
        \label{fig:synthetic_learned_progs}
    \end{subfigure}
    \begin{subfigure}[t]{0.17\linewidth}
    \centering
        \includegraphics[width=0.7\columnwidth]{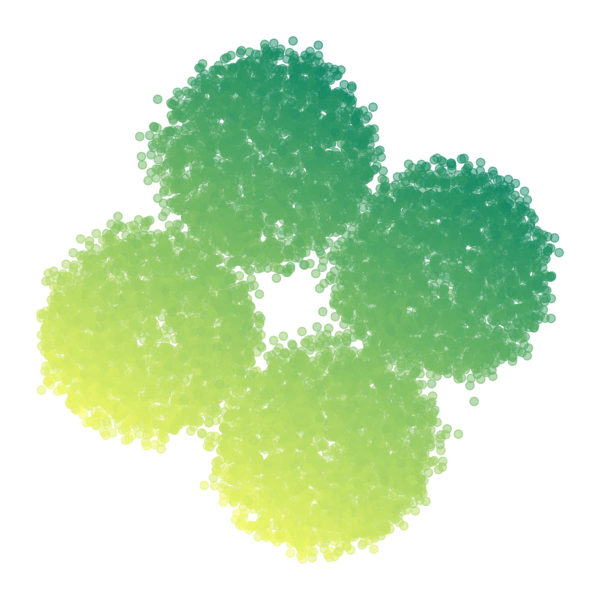}
        \includegraphics[width=0.7\columnwidth]{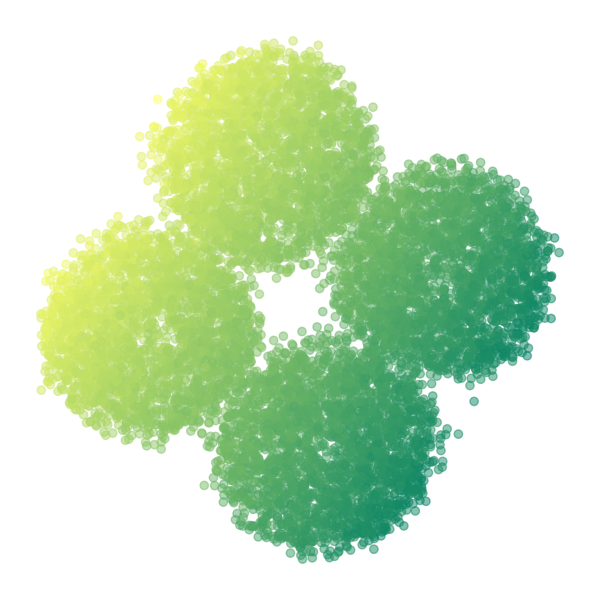}
        \caption{$\Zneural$, 0 programs}
        \label{fig:synthetic_tvae}
    \end{subfigure}
    \begin{subfigure}[t]{0.17\linewidth}
    \centering
        \includegraphics[width=0.7\columnwidth]{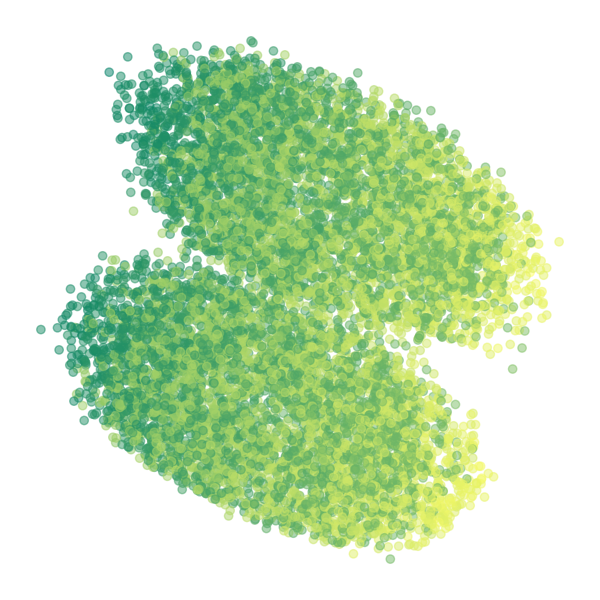}
        \includegraphics[width=0.7\columnwidth]{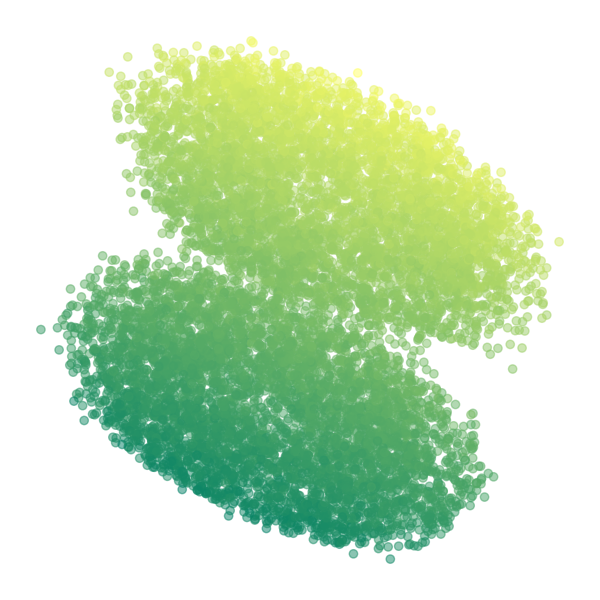}
        \caption{$\Zneural$, 1 program}
        \label{fig:synthetic_1prog}
    \end{subfigure}
    \begin{subfigure}[t]{0.17\linewidth}
    \centering
        \includegraphics[width=0.7\columnwidth]{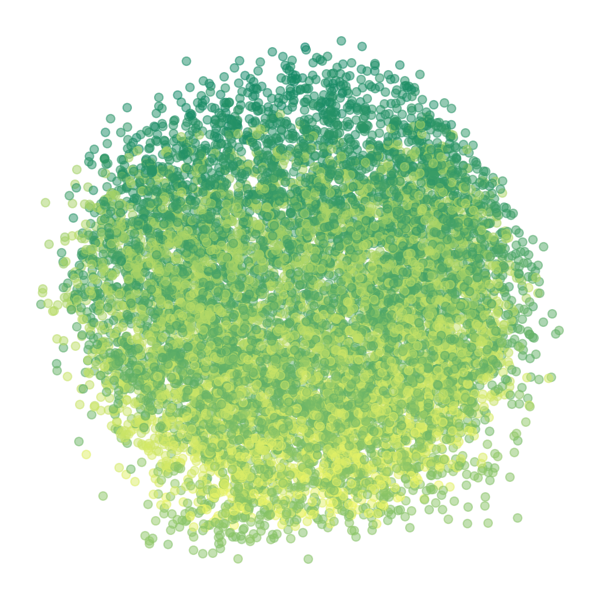}
        \includegraphics[width=0.7\columnwidth]{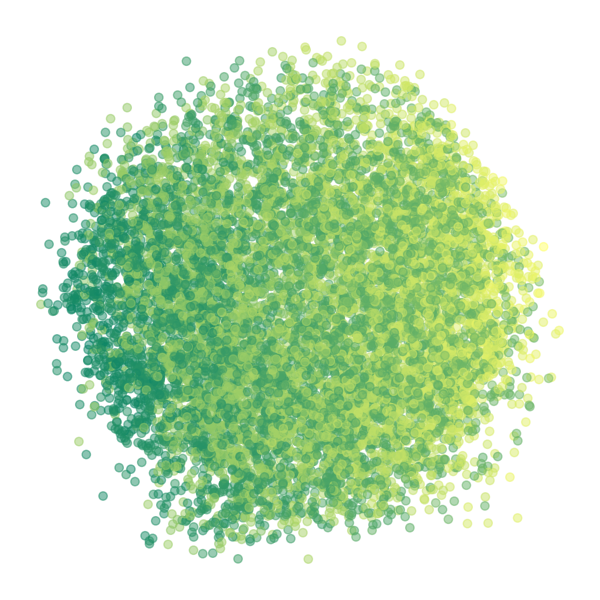}
        \caption{$\Zneural$, 2 programs}
        \label{fig:synthetic_2progs}
    \end{subfigure}
\vspace{-5pt}
\caption{ \textbf{Synthetic dataset experiments.} \textbf{(a)} Trajectories in synthetic training set. Initial/final positions are indicated in green/blue. Red lines delineate ground-truth classes, based on final positions. \textbf{(b)} $k=2$ learned binary programs using our algorithm. The first program (top) thresholds the final x-position while the second program (bottom) thresholds the final y-position. \textbf{(c, d, e)} Neural latent variables reduced to 2 dimensions. Top/bottom rows are colored by final x/y-positions respectively (green/yellow is positive/negative). \textbf{(c)} Clusters in TVAE neural latent space correspond to 4 ground-truth classes. \textbf{(d)} After learning the first program, the neural latent space contains clusters only based on the final y-position. \textbf{(e)} After learning the second program, all 4 ground-truth classes have been extracted as programs and the remaining neural latent space contains no clear clustering.}
\label{fig:synthetic}
\end{figure*}

\textbf{Metrics.} Unlike in the synthetic setting, we do not have ground truth programs in the real-world datasets.  We thus evaluate our programs quantitatively using (1) standard cluster metrics relative to human-defined labels, and (2) average precision for behavior classification when integrating our programs into downstream tasks. For cluster metrics, we use Purity~\citep{schutze2008introduction}, Normalized Mutual Information (NMI) \citep{zhang2006unsupervised}, and Rand Index (RI) \citep{rand1971objective}. We report the median of three runs. More details, including the standard deviation and the ELBO, are in the appendix. 

%%%
\subsection{Q1: Are the clusters created with our programs meaningful?}\label{sec:q1}
 
\begin{figure}[t]
    \centering
    \begin{subfigure}[b]{\linewidth}
    $$
    \mathbbm{1}_{[>-7]} \left[
    \begin{array}{l}
    \mapavg ~(\func~x_t.~ \\ 
    \ \ \mulc~(\mathit{ResidentSpeedAffine}_{[-6.3]; -8.3}(x_t), \\
    \ \ \ \  \mathit{NoseTailDistAffine}_{[.04];-9.1}(x_t))~x 
    \end{array}
    \right]
    $$
    %}
    \vspace{-0.02in}
    \caption{\textbf{Program learned using CalMS21 DSL 1}, resulting NMI 0.428. Since speed is positive, the first term is always negative. One cluster thus generally consists of trajectories where the mice are further apart, such that the second term is positive, and the negative product is less than the threshold. The other cluster generally occurs when the mice are close together, the second term is negative, and the product will be positive.}
    \label{fig:example_mouse}
    \end{subfigure}
    \\
    \begin{subfigure}[b]{\linewidth}
    %{\small 
    $$
    \mathbbm{1}_{[>-5.7]} \left[    
    \begin{array}{l}
    \mapavg ~(\func~x_t.~ \\ 
    \ \ \addc~(\mathit{ResidentAxisRatioAffine}_{[-8.0]; -7.1}(x_t), \\
    \ \ \ \ \mathit{BoundingBoxIOUAffine}_{[-16.6];5.9}(x_t))~x
    \end{array} 
    \right]
    $$
    %}
    \vspace{-0.02in}
    \caption{\textbf{Program learned using CalMS21 DSL 2}, resulting NMI 0.320. The axis ratio is the ratio of major axis length and minor axis length of an ellipse fitted to the mouse keypoints. The second term measures the bounding box overlap between mice, and is zero when the mice are far apart. It follows that one cluster generally contains trajectories when the mice has larger bounding box overlaps or if the resident axis ratio is large. The other cluster thus contains trajectories where the mice bounding boxes do not overlap, and resident body is compact.}
    \label{fig:example_mouse_2}    
    \end{subfigure}
        % \vspace{-0.25in}
    \caption{Learned programs on CalMS21. The subscripts represents the learned weights (in brackets) and biases (after the brackets) for the affine transformation followed by the bias.}
    \vspace{0.25in}
\end{figure}

\textbf{Synthetic dataset experiments.} Our synthetic dataset consists of trajectory data with 4 ground truth classes, corresponding to positive/negative x/y directions.
The goal is to learn symbolic programs that capture the ground-truth classes, while leaving the neural latent space to capture any residual information, such as the random initial velocity.
We visualize the 2 dimensions of the neural latent space of a TVAE along with 0, 1, and 2 learned programs in Figures \ref{fig:synthetic_tvae}, \ref{fig:synthetic_1prog} \& \ref{fig:synthetic_2progs}. The initial neural latent space of the TVAE contains 4 clusters corresponding to the 4 ground-truth classes in Figure \ref{fig:synthetic_tvae}. After our algorithm learns the first program that thresholds the final x-position, the resulting latent space in Figure \ref{fig:synthetic_1prog} captures the other factor of variation as 2 clusters corresponding to the final y-positions. Lastly, when our algorithm learns a second program that thresholds the final y-position, the resulting latent space in Figure \ref{fig:synthetic_2progs} no longer contains any clear clustering, showing that our approach has successfully extracted the 4 ground-truth classes.

\begin{table*}
  \begin{minipage}[c]{0.65\linewidth}
    \centering
\scalebox{0.9}{
 \begin{tabular}{c|c c c | c c c} 
        \toprule[0.2em]
         \multirow{2}{*}{Model} & 
         \multicolumn{3}{c|}{\textbf{CalMS21}} &
         \multicolumn{3}{c}{\textbf{Basketball}}\\ 
        & Purity & NMI & RI  & Purity & NMI & RI  \\ 
        \toprule[0.2em]
        Random assignment & .597 & .000 & .536 & .500 & .000 & .500 \\
        \hline
        TVAE  & .598 & .089 & .564 & .501 & .001 & .500 \\
        TVAE+KMeans loss & .605 & .118 & .573 & .501 & .001 & .500 \\ 
        JointVAE &  .597 & .019 & .537 & .560 & \textbf{.034} & .507 \\
        VQ-TVAE &  .601 & .124 & .588 & .572 & .016 & .511 \\
        Beta-TVAE & .616 & .115 & .589 & .565 & .013 & .509 \\
        \hline
        Ours (1 program)  & .706 & \textbf{.423} & \textbf{.694} & \textbf{.596} & .027 & \textbf{.518} \\
        Ours (2 programs)  & .725 & .320 & .648 & .561 & .033 & .507 \\ 
        Ours (3 programs)  & \textbf{.756} & .314 & .633 & .584 & .022 & .514 \\
        \toprule[0.2em]
 \end{tabular}
 }
  \end{minipage}
  \hfill
  \begin{minipage}[c]{0.35\linewidth}
    \caption{\textbf{Evaluating clusters from baseline and our neurosymbolic encoders on human-annotated labels.} Median purity, NMI, and RI on CalMS21 and Basketball compared to human-annotated labels (3 runs). Experiment hyperparameters are included in the appendix.  
    }
    \label{tab:cluster_metrics}
  \end{minipage}
  \vspace{0.2in}
\end{table*}

\textbf{Real-world datasets experiments.} We compare clusters produced by our neurosymbolic encoder with fully neural autoencoding baselines (Table~\ref{tab:cluster_metrics}), measured against human-annotated behaviors. For CalMS21, we observe that our method consistently outperforms the baselines in all three cluster metrics. 
We note that purity increases as the number of programs (thus clusters) increase, while NMI and RI decrease. This implies our method with two clusters best correspond to CalMS21 behaviors, but the other clusters found by our method may still be useful for domain experts. For Basketball, our method improves slightly with respect to purity, but is overall comparable with the baselines. 

\textbf{Qualitative interpretation of our clusters.} We further study the programs and clusters produced by our algorithm for the CalMS21 dataset, through a qualitative study with a  behavioral neuroscientist. Here, the behavioral neuroscientist analyzes the programmatic clusters produced from the symbolic representation of our neurosymbolic encoder for one, two, and three programs, resulting in two, four, and eight clusters respectively. The CalMS21 dataset is originally manually annotated with 4 classes corresponding to ``attack'', ``investigation'' (sniff), ``mount'', and ``other'' labels. ``Other'' corresponds to when no behaviors-of-interest is occuring, and is typically when the mice are not interacting.

In the single program case, our programs correspond to two discovered clusters. These clusters were classified by domain experts as referring to (1) when the mice are interacting and (2) when there are no interactions. They noted that this is based on distance between the mice, which is consistent with our program (Figure~\ref{fig:example_mouse}) using distance between nose of resident and tail of intruder. For two programs, there are a total of four discovered clusters, with two clusters each corresponding to no interaction and interaction. For the interaction clusters, the domain expert was further able to identify sniff tail behavior as one of the clusters. In this case, the programs found were based on intruder head body angle, resident nose and intruder tail distance, and resident nose and intruder nose distance. The domain expert found the three program case to be more difficult to interpret, but was able to identify clusters corresponding to sniff tail, resident exploration, interaction facing the same direction (ex: mounting), and interaction facing opposite directions (ex: face-to-face sniffing).

\subsection{Q2: How sensitive is our approach to the DSL?}\label{sec:q2}

\begin{table*}
\begin{center}
\scalebox{0.9}{
 \begin{tabular}{c|c c c | c c c| c c c } 
        \toprule[0.2em]
         \multirow{2}{*}{Model} & 
         \multicolumn{3}{c|}{\textbf{CalMS21 (DSL 1)}} &
         \multicolumn{3}{c|}{\textbf{CalMS21 (DSL 2)}} & \multicolumn{3}{c}{\textbf{CalMS21 (DSL 3)}}\\ 
        & Purity & NMI & RI  & Purity & NMI & RI & Purity & NMI & RI  \\  
        \toprule[0.2em]
        Ours (1 program)  & .706 & \textbf{.423} & \textbf{.694} & .689 & \textbf{.364} & \textbf{.681} & .649 & \textbf{.325} & .616 \\
        Ours (2 programs)  & \textbf{.725} & .320 & .648 & \textbf{.715} & .359 & .673 & \textbf{.664} & .324 & \textbf{.634} \\ 
        \toprule[0.2em]
 \end{tabular}}
\end{center}
\vspace{-0.1in}
\caption{ \textbf{Effect of varying DSLs on CalMS21 for neurosymbolic encoders.} Median purity, NMI, and RI on CalMS21 of our algorithms with DSLs selected by three domain experts compared to human-annotated labels (3 runs). DSL1 corresponds to Table~\ref{tab:cluster_metrics}. }
\label{tab:dsl_variations}
\vspace{-0.05in}
\end{table*}

\textbf{Choice of DSL.} 
To study the effect of DSL choices, we worked with three domain experts to construct three different DSLs used to learn our programmatic representations. These DSLs contained 8 to 10 different behavioral features for studying mouse social behavior on CalMS21, in addition to common sequential operations (Figure~\ref{fig:dsl}). A full list of features selected by domain experts are in the appendix. 

While there is some variability, our approach consistently outperforms the baselines that contain fully neural encoders for all three DSLs (Table~\ref{tab:dsl_variations}). Comparing some learned programs from two DSLs (Figures ~\ref{fig:example_mouse},~\ref{fig:example_mouse_2}), both contain a term that correlates with whether the mice are interacting (distance and bounding box overlap), and another term that correlates with resident speed (mice tends to be more stretched when they are moving quickly). 

\begin{table*}
  \begin{minipage}[c]{0.7\linewidth}
    \centering
\scalebox{0.9}{
 \begin{tabular}{c|c c c | c c c} 
        \toprule[0.2em]
         \multirow{2}{*}{Model} & 
         \multicolumn{3}{c|}{\textbf{CalMS21}} &
         \multicolumn{3}{c}{\textbf{Basketball}}\\ 
        & Purity & NMI & RI  & Purity & NMI & RI  \\ 
        \toprule[0.2em]
        TVAE  & .598 & .089 & .564 & .501 & .001 & .500 \\
        TVAE (w/ features) & .597 & .103 & .570 & .565 & .012 & .508 \\ 
        \hline
        VQ-TVAE &  .601 & .124 & .588 & .571 & .016 & .511\\
        VQ-TVAE (w/ features) &  .608 & .114 & .601 & .525 & .002 & .501 \\
        \hline
        Beta-TVAE & .616 & .115 & .589 & .566 & .013 & .509 \\
        Beta-TVAE (w/ features) & .612 & .096 & .571 & .563 & .011 & .508 \\        
        \toprule[0.2em]
 \end{tabular}}
  \end{minipage}
  \hfill
  \begin{minipage}[c]{0.3\linewidth}
    \vspace{0.1in}
    \caption{\textbf{Effect of encoding DSL features into baselines.} Median purity, NMI, and RI on CalMS21 and Basketball compared to human-annotated labels (3 runs) for baseline with trajectory inputs only, and baseline with trajectory features added.
    }
    \label{tab:baseline_features}
  \end{minipage}
  \vspace{-0.05in}
\end{table*}

\textbf{DSL features as input.} Lastly, we experiment with using the same DSL features introduced by domain experts as additional features for input trajectories instead (Table~\ref{tab:baseline_features}). For both CalMS21 and Basketball, the baselines using the additional features have comparable performance to using input trajectory data alone. In contrast, by using the features more explicitly as part of the DSL in our neurosymbolic encoders, we are able to produce clusters with a better separation between behavior classes based on cluster metrics (see Table \ref{tab:cluster_metrics}).

\subsection{Q3: Are the programs useful for downstream tasks?}\label{sec:q3}

We apply our programs to frame-level behavior classification~\citep{segalin2020mouse,eyjolfsdottir2016learning,burgos2012social}, where the goal is to automatically quantify behavior based on expert annotations. We are motivated by the observation that manual behavior annotation is time-consuming and expensive~\citep{anderson2014toward}, often being a bottleneck in the analysis workflow. Our unsupervised programs have the potential to reduce annotation effort and help accelerate behavioral studies, through the task programming framework. Task programming~\citep{sun2020task} uses hand-crafted programs as self-supervision to improve behavior classification data efficiency; however, hand designing programs still requires human effort. Here, we show that unsupervised programs learned using our neurosymbolic encoders performs comparably to expert-designed programs on CalMS21.

We integrate the learned programs from our neurosymbolic encoder into the task programming framework (i.e., use them as a source of self-supervision instead of the expert-crafted programs), and compare to the classification performance using expert programs (Figure~\ref{fig:task_programming}). The classification performance is computed using Mean Average Precision on the behaviors-of-interest in CalMS21 (attack, investigation, mount).
Using only one program found using our approach, we are able to achieve comparable performance to 10 expert-written programs on the behavior classification task studied in~\cite{sun2020task}. 
Importantly, we note that we automatically learned the self-supervision tasks from a DSL, instead of hand-crafting them as in~\cite{sun2020task}.
This demonstrates that programs found by our approach can be applied effectively to downstream behavior analysis tasks such as task programming. 

\begin{figure}
    \centering
        \centering
        \hspace{-0.15in}
        \includegraphics[width=0.7\linewidth]{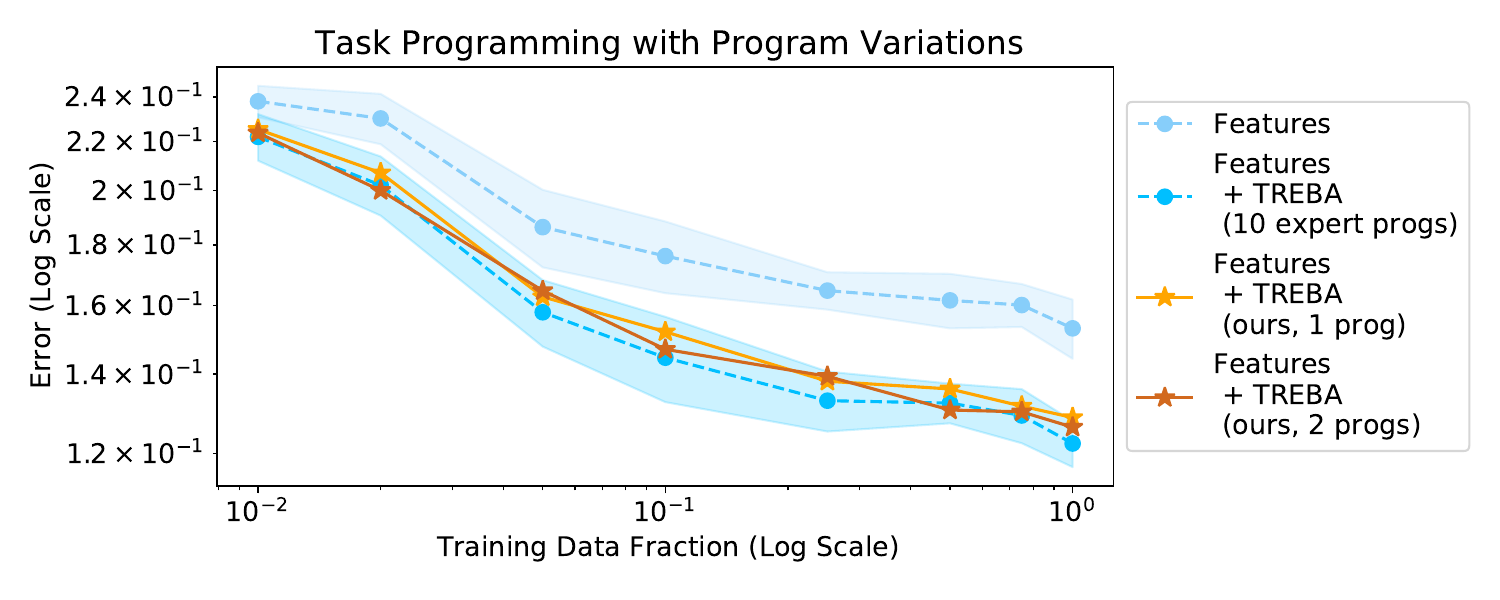}
        \vspace{-0.15in}
    \caption{ \textbf{Applying symbolic encoders for self-supervision.} ``Features'' is baseline w/o self-supervision. ``TREBA'' is a self-supervised approach in the Task Programming paradigm \citep{sun2020task}, using either expert-crafted programs or our symbolic encoders as the weak-supervision rules. The shaded region is std dev over 9 repeats. The std dev for our approach (not shown) is comparable. Based on~\cite{sun2020task}, the error is computed using $1.0 - \text{Mean Average Precision}$. 
    }
    \label{fig:task_programming}
    \vspace{0.05in}
\end{figure}

\section{Other Related Work}
\label{sec:related}

%%%
\textbf{Interpretable latent variable models.} 
Latent representations, especially those that are human-interpretable, can help us understand the structure of data. These models may learn disentangled factors~\citep{higgins2016beta,chen2016infogan,ma2020disentangled} or semantically meaningful clusters~\citep{ma2019learning} using unsupervised learning approaches. These approaches are often grounded in the VAE framework~\citep{kingma2013auto}. 

Some of these approaches, such as JointVAE~\citep{dupont2018learning}, Discrete VAE~\citep{rolfe2016discrete}, Guided-VAE~\citep{ding2020guided}, and VQ-VAE~\citep{oord2017neural}, learn discrete latent representations (in particular,  JointVAE learns a combined discrete-continuous representation, just like our approach).
However, these approaches use fully neural encoders. To our knowledge, our work is the first to propose neurosymbolic encoders, where the symbolic component produces a symbolic program that produces an low-dimensional encoding of the input data.

%%%
\textbf{Neurosymbolic programming.}
Neurosymbolic programming \citep{chaudhuri2021neurosymbolic} has seen much activity in the recent past.
Existing approaches here are often trained in a supervised fashion~\citep{gulwani2011automating,wang2017program,shah2020learning,cui2021differentiable}, or within a (generative) policy learning context with an explicit reward function~\citep{chen2017towards,verma2018programmatically,verma2019imitation,bastani2018verifiable,inala2019synthesizing,feinman2020learning,trivedi2021learning}.  
Prior work on unsupervised program synthesis has mostly addressed generative modeling, i.e., the synthesis of programs that can generate the training data \citep{tian2018learning, ellis2017learning,feinman2020learning}. This task is analogous to learning a symbolic decoder rather than a symbolic encoder. Studying how to incorporate such methods into our framework can be an interesting future direction.

%%%
\textbf{Representation learning for behavior analysis.}
Representation learning has been applied to a variety of downstream tasks for behavior analysis, such as discovering behavior motifs~\citep{berman2014mapping, singh2021mining}, identifying internal states~\citep{calhoun2019unsupervised}, and improving sample-efficiency~\citep{sun2020task}. Studies in this area have used methods such as VAE~\citep{kingma2013auto}, AR-HMM~\citep{wiltschko2015mapping}, forecasting or predicting future behaviors~\citep{liang2020learning,gao2020vectornet}, and  Umap~\citep{mcinnes2018umap} to better understand the latent structure of behavior. Similar to a few other representation learning methods~\citep{luxem2020identifying,sun2020task}, we also use an encoder-decoder setup on trajectory data. However, our work learns a neurosymbolic encoder whereas existing works in this area have fully neural encoders. Our work can aid behavior analysis by learning more interpretable latent representations and can be applied to downstream tasks, such as behavior classification.

% \vspace{-0.1in}

\section{Discussion}~\label{sec:conclusion}
\vspace{-0.2in}

We present a novel approach for unsupervised learning of neurosymbolic encoders. Our approach integrates the VAE framework with program synthesis and results in a learned representation with both neural and symbolic components. Experiments on trajectory data from behavior analysis demonstrate that our programmatic descriptions of the latent space result in more meaningful clusters relative to human-defined behaviors, compared to purely neural encoders. Additionally, we show the practicality of our approach by applying our learned programs to achieve comparable performance to expert-constructed tasks in a self-supervised learning approach for behavior classification. 

\textbf{Problem Scope.} We explore unsupervised learning of neurosymbolic encoders for the first time, and here, our neurosymbolic encoders tackle domains consisting of lower dimensional spatiotemporal data. These types of domains covers a wide range of application areas, from behavioral data (animal behavior and sports analytics in our experiments), to control systems for rigid-body systems, to biomarkers or socioeconomic markers. In many of these domains, there are existing domain expertise that can be leveraged to create the DSL for our neurosymbolic encoders. For example, we use the behavioral features from~\cite{segalin2020mouse} in our work. 
One direct application of learning semantically meaningful programs is that it can be used to improve learning pipelines, such as task programming, as we have demonstrated. 

\textbf{Limitations.} One limitation of our current approach is scalability of the program search process. While our program search is parallelizable, such that learning additional programs would not incur significant additional time, the symbolic encoder update does increase the runtime over a purely neural solution. 
Here, we have explored our approach on settings where shorter programs are beneficial.
Future work have the potential to further expand the applications of these models to larger, more complex systems. 
Furthermore, our approach requires programs that are differentiable with respect to its parameters.
We note that there are increasingly more differentiable DSLs, such as~\cite{shah2020learning,cui2021differentiable,valkov2018houdini,gaunt2016terpret,bunel2016adaptive}, and there are commonly-adopted ways to make differentiable approximations to more established non-differentiable DSLs (for example, in~\cite{shah2020learning}, the authors use a smooth differentiable approximation of the non-differentiable if-then-else statement). 
These common challenges in using neurosymbolic learning in science is further discussed in~\cite{sun2022neurosymbolic}.

\textbf{Future Directions.} There are many future directions to explore for neurosymbolic encoders based on our work. 
Scalability is one important area as discussed above. 
Another direction is to extend this work to other domains such as image and text data, in order to learn interpretable symbolic latent representations. Neurosymbolic encoders on images would require a DSL for pixel data as well as architecture changes, such as using convolutional VAEs. 
Furthermore, one can improve upon our greedy approach in Algorithm \ref{alg:mutliple_alg} for finding the optimal set of symbolic programs, e.g. by performing local coordinate ascent in program space, similar to algorithms for large-scale neighborhood search \citep{ahuja2002survey}. Lastly, while practically-oriented extensions of VAEs such as our own have yielded great practical benefit, they often lead to sub-optimal results from a pure likelihood (or ELBO) perspective. One final direction is to rigorously formulate a learning objective from the ground up that formally encapsulates practically-oriented extensions of VAEs.

\textbf{Acknowledgements.} The authors are grateful to the anonymous reviewers for their helpful comments. 
This work was funded in part by NSF \#1918865, and a gift from Amazon.

\bibliography{main}

\begin{thebibliography}{66}
\providecommand{\natexlab}[1]{#1}
\providecommand{\url}[1]{\texttt{#1}}
\expandafter\ifx\csname urlstyle\endcsname\relax
  \providecommand{\doi}[1]{doi: #1}\else
  \providecommand{\doi}{doi: \begingroup \urlstyle{rm}\Url}\fi

\bibitem[Ahuja et~al.(2002)Ahuja, Ergun, Orlin, and Punnen]{ahuja2002survey}
Ravindra~K Ahuja, {\"O}zlem Ergun, James~B Orlin, and Abraham~P Punnen.
\newblock A survey of very large-scale neighborhood search techniques.
\newblock \emph{Discrete Applied Mathematics}, 123\penalty0 (1-3):\penalty0
  75--102, 2002.

\bibitem[Anderson \& Perona(2014)Anderson and Perona]{anderson2014toward}
David~J Anderson and Pietro Perona.
\newblock Toward a science of computational ethology.
\newblock \emph{Neuron}, 84\penalty0 (1):\penalty0 18--31, 2014.

\bibitem[Bastani et~al.(2018)Bastani, Pu, and
  Solar-Lezama]{bastani2018verifiable}
Osbert Bastani, Yewen Pu, and Armando Solar-Lezama.
\newblock Verifiable reinforcement learning via policy extraction.
\newblock In \emph{Advances in Neural Information Processing Systems}, 2018.

\bibitem[Berman et~al.(2014)Berman, Choi, Bialek, and
  Shaevitz]{berman2014mapping}
Gordon~J Berman, Daniel~M Choi, William Bialek, and Joshua~W Shaevitz.
\newblock Mapping the stereotyped behaviour of freely moving fruit flies.
\newblock \emph{Journal of The Royal Society Interface}, 11\penalty0
  (99):\penalty0 20140672, 2014.

\bibitem[Bowman et~al.(2015)Bowman, Vilnis, Vinyals, Dai, J{\'{o}}zefowicz, and
  Bengio]{bowman2015generating}
Samuel~R. Bowman, Luke Vilnis, Oriol Vinyals, Andrew~M. Dai, Rafal
  J{\'{o}}zefowicz, and Samy Bengio.
\newblock Generating sentences from a continuous space.
\newblock \emph{arXiv preprint arXiv:1511.06349}, 2015.

\bibitem[Bunel et~al.(2016)Bunel, Desmaison, Mudigonda, Kohli, and
  Torr]{bunel2016adaptive}
Rudy~R Bunel, Alban Desmaison, Pawan~K Mudigonda, Pushmeet Kohli, and Philip
  Torr.
\newblock Adaptive neural compilation.
\newblock \emph{Advances in Neural Information Processing Systems}, 29, 2016.

\bibitem[Burgess et~al.(2017)Burgess, Higgins, Pal, Matthey, Watters,
  Desjardins, and Lerchner]{burgess2018understanding}
Christopher~P. Burgess, Irina Higgins, Arka Pal, Loic Matthey, Nick Watters,
  Guillaume Desjardins, and Alexander Lerchner.
\newblock Understanding disentangling in $\beta$-vae.
\newblock In \emph{Neural Information Processing Systems Disentanglement
  Workshop}, 2017.

\bibitem[Burgos-Artizzu et~al.(2012)Burgos-Artizzu, Doll{\'a}r, Lin, Anderson,
  and Perona]{burgos2012social}
Xavier~P Burgos-Artizzu, Piotr Doll{\'a}r, Dayu Lin, David~J Anderson, and
  Pietro Perona.
\newblock Social behavior recognition in continuous video.
\newblock In \emph{2012 IEEE Conference on Computer Vision and Pattern
  Recognition}, pp.\  1322--1329. IEEE, 2012.

\bibitem[Calhoun et~al.(2019)Calhoun, Pillow, and
  Murthy]{calhoun2019unsupervised}
Adam~J Calhoun, Jonathan~W Pillow, and Mala Murthy.
\newblock Unsupervised identification of the internal states that shape natural
  behavior.
\newblock \emph{Nature neuroscience}, 22\penalty0 (12):\penalty0 2040--2049,
  2019.

\bibitem[Chaudhuri et~al.(2021)Chaudhuri, Ellis, Polozov, Singh, Solar-Lezama,
  Yue, et~al.]{chaudhuri2021neurosymbolic}
Swarat Chaudhuri, Kevin Ellis, Oleksandr Polozov, Rishabh Singh, Armando
  Solar-Lezama, Yisong Yue, et~al.
\newblock Neurosymbolic programming.
\newblock \emph{Foundations and Trends{\textregistered} in Programming
  Languages}, 7\penalty0 (3):\penalty0 158--243, 2021.

\bibitem[Chen et~al.(2016{\natexlab{a}})Chen, Duan, Houthooft, Schulman,
  Sutskever, and Abbeel]{chen2016infogan}
Xi~Chen, Yan Duan, Rein Houthooft, John Schulman, Ilya Sutskever, and Pieter
  Abbeel.
\newblock Infogan: Interpretable representation learning by information
  maximizing generative adversarial nets.
\newblock \emph{arXiv preprint arXiv:1606.03657}, 2016{\natexlab{a}}.

\bibitem[Chen et~al.(2016{\natexlab{b}})Chen, Kingma, Salimans, Duan, Dhariwal,
  Schulman, Sutskever, and Abbeel]{chen2016variational}
Xi~Chen, Diederik~P Kingma, Tim Salimans, Yan Duan, Prafulla Dhariwal, John
  Schulman, Ilya Sutskever, and Pieter Abbeel.
\newblock Variational lossy autoencoder.
\newblock \emph{arXiv preprint arXiv:1611.02731}, 2016{\natexlab{b}}.

\bibitem[Chen et~al.(2018)Chen, Liu, and Song]{chen2017towards}
Xinyun Chen, Chang Liu, and Dawn Song.
\newblock Towards synthesizing complex programs from input-output examples.
\newblock In \emph{International Conference on Learning Representations}, 2018.

\bibitem[Co-Reyes et~al.(2018)Co-Reyes, Liu, Gupta, Eysenbach, Abbeel, and
  Levine]{co2018self}
John Co-Reyes, YuXuan Liu, Abhishek Gupta, Benjamin Eysenbach, Pieter Abbeel,
  and Sergey Levine.
\newblock Self-consistent trajectory autoencoder: Hierarchical reinforcement
  learning with trajectory embeddings.
\newblock In \emph{International Conference on Machine Learning}, 2018.

\bibitem[Creswell et~al.(2017)Creswell, Mohamied, Sengupta, and
  Bharath]{creswell2017adversarial}
Antonia Creswell, Yumnah Mohamied, Biswa Sengupta, and Anil~A Bharath.
\newblock Adversarial information factorization.
\newblock \emph{arXiv preprint arXiv:1711.05175}, 2017.

\bibitem[Cui \& Zhu(2021)Cui and Zhu]{cui2021differentiable}
Guofeng Cui and He~Zhu.
\newblock Differentiable synthesis of program architectures.
\newblock \emph{Advances in Neural Information Processing Systems},
  34:\penalty0 11123--11135, 2021.

\bibitem[Deng et~al.(2017)Deng, Navarathna, Carr, Mandt, Yue, Matthews, and
  Mori]{deng2017factorized}
Zhiwei Deng, Rajitha Navarathna, Peter Carr, Stephan Mandt, Yisong Yue, Iain
  Matthews, and Greg Mori.
\newblock Factorized variational autoencoders for modeling audience reactions
  to movies.
\newblock In \emph{IEEE conference on computer vision and pattern recognition},
  2017.

\bibitem[Ding et~al.(2020)Ding, Xu, Xu, Parmar, Yang, Welling, and
  Tu]{ding2020guided}
Zheng Ding, Yifan Xu, Weijian Xu, Gaurav Parmar, Yang Yang, Max Welling, and
  Zhuowen Tu.
\newblock Guided variational autoencoder for disentanglement learning.
\newblock In \emph{Proceedings of the IEEE/CVF Conference on Computer Vision
  and Pattern Recognition}, pp.\  7920--7929, 2020.

\bibitem[Dupont(2018)]{dupont2018learning}
Emilien Dupont.
\newblock Learning disentangled joint continuous and discrete representations.
\newblock In \emph{Neural Information Processing Systems}, 2018.

\bibitem[Ellis et~al.(2018)Ellis, Ritchie, Solar-Lezama, and
  Tenenbaum]{ellis2017learning}
Kevin Ellis, Daniel Ritchie, Armando Solar-Lezama, and Joshua~B Tenenbaum.
\newblock Learning to infer graphics programs from hand-drawn images.
\newblock In \emph{Advances in Neural Information Processing Systems}, 2018.

\bibitem[Eyjolfsdottir et~al.(2016)Eyjolfsdottir, Branson, Yue, and
  Perona]{eyjolfsdottir2016learning}
Eyrun Eyjolfsdottir, Kristin Branson, Yisong Yue, and Pietro Perona.
\newblock Learning recurrent representations for hierarchical behavior
  modeling.
\newblock \emph{arXiv preprint arXiv:1611.00094}, 2016.

\bibitem[Feinman \& Lake(2020)Feinman and Lake]{feinman2020learning}
Reuben Feinman and Brenden~M Lake.
\newblock Learning task-general representations with generative neuro-symbolic
  modeling.
\newblock In \emph{International Conference on Learning Representations}, 2020.

\bibitem[Gao et~al.(2020)Gao, Sun, Zhao, Shen, Anguelov, Li, and
  Schmid]{gao2020vectornet}
Jiyang Gao, Chen Sun, Hang Zhao, Yi~Shen, Dragomir Anguelov, Congcong Li, and
  Cordelia Schmid.
\newblock Vectornet: Encoding hd maps and agent dynamics from vectorized
  representation.
\newblock In \emph{Proceedings of the IEEE/CVF Conference on Computer Vision
  and Pattern Recognition}, pp.\  11525--11533, 2020.

\bibitem[Gaunt et~al.(2016)Gaunt, Brockschmidt, Singh, Kushman, Kohli, Taylor,
  and Tarlow]{gaunt2016terpret}
Alexander~L Gaunt, Marc Brockschmidt, Rishabh Singh, Nate Kushman, Pushmeet
  Kohli, Jonathan Taylor, and Daniel Tarlow.
\newblock Terpret: A probabilistic programming language for program induction.
\newblock \emph{arXiv preprint arXiv:1608.04428}, 2016.

\bibitem[Gulwani(2011)]{gulwani2011automating}
Sumit Gulwani.
\newblock Automating string processing in spreadsheets using input-output
  examples.
\newblock \emph{ACM Sigplan Notices}, 46\penalty0 (1):\penalty0 317--330, 2011.

\bibitem[Higgins et~al.(2016)Higgins, Matthey, Pal, Burgess, Glorot, Botvinick,
  Mohamed, and Lerchner]{higgins2016beta}
Irina Higgins, Loic Matthey, Arka Pal, Christopher Burgess, Xavier Glorot,
  Matthew Botvinick, Shakir Mohamed, and Alexander Lerchner.
\newblock beta-vae: Learning basic visual concepts with a constrained
  variational framework.
\newblock In \emph{International Conference on Learning Representations}, 2016.

\bibitem[Hsu \& Yttri(2020)Hsu and Yttri]{hsu2020b}
Alexander~I Hsu and Eric~A Yttri.
\newblock B-soid: An open source unsupervised algorithm for discovery of
  spontaneous behaviors.
\newblock \emph{bioRxiv}, pp.\  770271, 2020.

\bibitem[Hu et~al.(2017)Hu, Yang, Liang, Salakhutdinov, and Xing]{hu2017toward}
Zhiting Hu, Zichao Yang, Xiaodan Liang, Ruslan Salakhutdinov, and Eric~P Xing.
\newblock Toward controlled generation of text.
\newblock In \emph{International Conference on Machine Learning}, 2017.

\bibitem[Inala et~al.(2020)Inala, Bastani, Tavares, and
  Solar-Lezama]{inala2019synthesizing}
Jeevana~Priya Inala, Osbert Bastani, Zenna Tavares, and Armando Solar-Lezama.
\newblock Synthesizing programmatic policies that inductively generalize.
\newblock In \emph{International Conference on Learning Representations}, 2020.

\bibitem[Jang et~al.(2017)Jang, Gu, and Poole]{jang2017categorical}
Eric Jang, Shixiang Gu, and Ben Poole.
\newblock Categorical reparameterization with gumbel-softmax.
\newblock \emph{arXiv preprint arXiv:1611.01144}, 2017.

\bibitem[Johnson et~al.(2016)Johnson, Duvenaud, Wiltschko, Datta, and
  Adams]{johnson2016composing}
Matthew~J Johnson, David Duvenaud, Alexander~B Wiltschko, Sandeep~R Datta, and
  Ryan~P Adams.
\newblock Composing graphical models with neural networks for structured
  representations and fast inference.
\newblock In \emph{Advances in Neural Information Processing Systems}, 2016.

\bibitem[Kaiser et~al.(2018)Kaiser, Bengio, Roy, Vaswani, Parmar, Uszkoreit,
  and Shazeer]{kaiser2018fast}
Lukasz Kaiser, Samy Bengio, Aurko Roy, Ashish Vaswani, Niki Parmar, Jakob
  Uszkoreit, and Noam Shazeer.
\newblock Fast decoding in sequence models using discrete latent variables.
\newblock In \emph{International Conference on Machine Learning}, 2018.

\bibitem[Kim \& Mnih(2018)Kim and Mnih]{kim2018disentangling}
Hyunjik Kim and Andriy Mnih.
\newblock Disentangling by factorising.
\newblock In \emph{International Conference on Machine Learning}, 2018.

\bibitem[Kingma \& Ba(2014)Kingma and Ba]{kingma2014adam}
Diederik~P Kingma and Jimmy Ba.
\newblock Adam: A method for stochastic optimization.
\newblock \emph{arXiv preprint arXiv:1412.6980}, 2014.

\bibitem[Kingma \& Welling(2014)Kingma and Welling]{kingma2013auto}
Diederik~P Kingma and Max Welling.
\newblock Auto-encoding variational bayes.
\newblock In \emph{International Conference on Learning Representations}, 2014.

\bibitem[Kingma et~al.(2014)Kingma, Rezende, Mohamed, and
  Welling]{kingma2014semi}
Diederik~P Kingma, Danilo~J Rezende, Shakir Mohamed, and Max Welling.
\newblock Semi-supervised learning with deep generative models.
\newblock \emph{arXiv preprint arXiv:1406.5298}, 2014.

\bibitem[Liang et~al.(2020)Liang, Yang, Hu, Chen, Liao, Feng, and
  Urtasun]{liang2020learning}
Ming Liang, Bin Yang, Rui Hu, Yun Chen, Renjie Liao, Song Feng, and Raquel
  Urtasun.
\newblock Learning lane graph representations for motion forecasting.
\newblock In \emph{European Conference on Computer Vision}, pp.\  541--556.
  Springer, 2020.

\bibitem[Lloyd(1982)]{lloyd1982least}
Stuart Lloyd.
\newblock Least squares quantization in pcm.
\newblock \emph{IEEE transactions on information theory}, 28\penalty0
  (2):\penalty0 129--137, 1982.

\bibitem[Luxem et~al.(2020)Luxem, Fuhrmann, K{\"u}rsch, Remy, and
  Bauer]{luxem2020identifying}
Kevin Luxem, Falko Fuhrmann, Johannes K{\"u}rsch, Stefan Remy, and Pavol Bauer.
\newblock Identifying behavioral structure from deep variational embeddings of
  animal motion.
\newblock \emph{bioRxiv}, 2020.

\bibitem[Ma et~al.(2020)Ma, Zhou, Yang, Cui, Wang, and Zhu]{ma2020disentangled}
Jianxin Ma, Chang Zhou, Hongxia Yang, Peng Cui, Xin Wang, and Wenwu Zhu.
\newblock Disentangled self-supervision in sequential recommenders.
\newblock In \emph{Proceedings of the 26th ACM SIGKDD International Conference
  on Knowledge Discovery \& Data Mining}, pp.\  483--491, 2020.

\bibitem[Ma et~al.(2019)Ma, Zheng, Li, and Cottrell]{ma2019learning}
Qianli Ma, Jiawei Zheng, Sen Li, and Gary~W Cottrell.
\newblock Learning representations for time series clustering.
\newblock \emph{Advances in neural information processing systems},
  32:\penalty0 3781--3791, 2019.

\bibitem[McInnes et~al.(2018)McInnes, Healy, and Melville]{mcinnes2018umap}
Leland McInnes, John Healy, and James Melville.
\newblock Umap: Uniform manifold approximation and projection for dimension
  reduction.
\newblock \emph{arXiv preprint arXiv:1802.03426}, 2018.

\bibitem[Mnih \& Gregor(2014)Mnih and Gregor]{mnih2014neural}
Andriy Mnih and Karol Gregor.
\newblock Neural variational inference and learning in belief networks.
\newblock In \emph{International Conference on Machine Learning}, 2014.

\bibitem[Oord et~al.(2017)Oord, Vinyals, and Kavukcuoglu]{oord2017neural}
Aaron van~den Oord, Oriol Vinyals, and Koray Kavukcuoglu.
\newblock Neural discrete representation learning.
\newblock In \emph{Advances in Neural Information Processing Systems}, 2017.

\bibitem[Rand(1971)]{rand1971objective}
William~M Rand.
\newblock Objective criteria for the evaluation of clustering methods.
\newblock \emph{Journal of the American Statistical association}, 66\penalty0
  (336):\penalty0 846--850, 1971.

\bibitem[Rolfe(2016)]{rolfe2016discrete}
Jason~Tyler Rolfe.
\newblock Discrete variational autoencoders.
\newblock \emph{arXiv preprint arXiv:1609.02200}, 2016.

\bibitem[Sch{\"u}tze et~al.(2008)Sch{\"u}tze, Manning, and
  Raghavan]{schutze2008introduction}
Hinrich Sch{\"u}tze, Christopher~D Manning, and Prabhakar Raghavan.
\newblock \emph{Introduction to information retrieval}, volume~39.
\newblock Cambridge University Press Cambridge, 2008.

\bibitem[Segalin et~al.(2020)Segalin, Williams, Karigo, Hui, Zelikowsky, Sun,
  Perona, Anderson, and Kennedy]{segalin2020mouse}
Cristina Segalin, Jalani Williams, Tomomi Karigo, May Hui, Moriel Zelikowsky,
  Jennifer~J Sun, Pietro Perona, David~J Anderson, and Ann Kennedy.
\newblock The mouse action recognition system (mars): a software pipeline for
  automated analysis of social behaviors in mice.
\newblock \emph{bioRxiv}, 2020.

\bibitem[Shah et~al.(2020)Shah, Zhan, Sun, Verma, Yue, and
  Chaudhuri]{shah2020learning}
Ameesh Shah, Eric Zhan, Jennifer~J Sun, Abhinav Verma, Yisong Yue, and Swarat
  Chaudhuri.
\newblock Learning differentiable programs with admissible neural heuristics.
\newblock In \emph{Neural Information Processing Systems}, 2020.

\bibitem[Singh et~al.(2021)Singh, Peterson, Rao, and Brunton]{singh2021mining}
Satpreet~H Singh, Steven~M Peterson, Rajesh~PN Rao, and Bingni~W Brunton.
\newblock Mining naturalistic human behaviors in long-term video and neural
  recordings.
\newblock \emph{Journal of Neuroscience Methods}, 2021.

\bibitem[Sun et~al.(2021{\natexlab{a}})Sun, Karigo, Chakraborty, Mohanty,
  Anderson, Perona, Yue, and Kennedy]{sun2021multi}
Jennifer~J Sun, Tomomi Karigo, Dipam Chakraborty, Sharada~P Mohanty, David~J
  Anderson, Pietro Perona, Yisong Yue, and Ann Kennedy.
\newblock The multi-agent behavior dataset: Mouse dyadic social interactions.
\newblock \emph{arXiv preprint arXiv:2104.02710}, 2021{\natexlab{a}}.

\bibitem[Sun et~al.(2021{\natexlab{b}})Sun, Kennedy, Zhan, Anderson, Yue, and
  Perona]{sun2020task}
Jennifer~J Sun, Ann Kennedy, Eric Zhan, David~J Anderson, Yisong Yue, and
  Pietro Perona.
\newblock Task programming: Learning data efficient behavior representations.
\newblock In \emph{Conference on Computer Vision and Pattern Recognition},
  2021{\natexlab{b}}.

\bibitem[Sun et~al.(2022)Sun, Tjandrasuwita, Sehgal, Solar-Lezama, Chaudhuri,
  Yue, and Costilla-Reyes]{sun2022neurosymbolic}
Jennifer~J Sun, Megan Tjandrasuwita, Atharva Sehgal, Armando Solar-Lezama,
  Swarat Chaudhuri, Yisong Yue, and Omar Costilla-Reyes.
\newblock Neurosymbolic programming for science.
\newblock \emph{arXiv preprint arXiv:2210.05050}, 2022.

\bibitem[Tian et~al.(2018)Tian, Luo, Sun, Ellis, Freeman, Tenenbaum, and
  Wu]{tian2018learning}
Yonglong Tian, Andrew Luo, Xingyuan Sun, Kevin Ellis, William~T Freeman,
  Joshua~B Tenenbaum, and Jiajun Wu.
\newblock Learning to infer and execute 3d shape programs.
\newblock In \emph{International Conference on Learning Representations}, 2018.

\bibitem[Trivedi et~al.(2021)Trivedi, Zhang, Sun, and Lim]{trivedi2021learning}
Dweep Trivedi, Jesse Zhang, Shao-Hua Sun, and Joseph~J Lim.
\newblock Learning to synthesize programs as interpretable and generalizable
  policies.
\newblock \emph{Advances in neural information processing systems},
  34:\penalty0 25146--25163, 2021.

\bibitem[Vahdat \& Kautz(2020)Vahdat and Kautz]{vahdat2020nvae}
Arash Vahdat and Jan Kautz.
\newblock Nvae: A deep hierarchical variational autoencoder.
\newblock In \emph{Advances in Neural Information Processing Systems}, 2020.

\bibitem[Valkov et~al.(2018)Valkov, Chaudhari, Srivastava, Sutton, and
  Chaudhuri]{valkov2018houdini}
Lazar Valkov, Dipak Chaudhari, Akash Srivastava, Charles Sutton, and Swarat
  Chaudhuri.
\newblock Houdini: Lifelong learning as program synthesis.
\newblock In \emph{Advances in neural information processing systems}, 2018.

\bibitem[Verma et~al.(2018)Verma, Murali, Singh, Kohli, and
  Chaudhuri]{verma2018programmatically}
Abhinav Verma, Vijayaraghavan Murali, Rishabh Singh, Pushmeet Kohli, and Swarat
  Chaudhuri.
\newblock Programmatically interpretable reinforcement learning.
\newblock In \emph{International Conference on Machine Learning}, 2018.

\bibitem[Verma et~al.(2019)Verma, Le, Yue, and Chaudhuri]{verma2019imitation}
Abhinav Verma, Hoang~M Le, Yisong Yue, and Swarat Chaudhuri.
\newblock Imitation-projected programmatic reinforcement learning.
\newblock In \emph{Advances in Neural Information Processing Systems}, 2019.

\bibitem[Wang et~al.(2017)Wang, Dillig, and Singh]{wang2017program}
Xinyu Wang, Isil Dillig, and Rishabh Singh.
\newblock Program synthesis using abstraction refinement.
\newblock \emph{Proceedings of the ACM on Programming Languages}, 2017.

\bibitem[Wiltschko et~al.(2015)Wiltschko, Johnson, Iurilli, Peterson, Katon,
  Pashkovski, Abraira, Adams, and Datta]{wiltschko2015mapping}
Alexander~B Wiltschko, Matthew~J Johnson, Giuliano Iurilli, Ralph~E Peterson,
  Jesse~M Katon, Stan~L Pashkovski, Victoria~E Abraira, Ryan~P Adams, and
  Sandeep~Robert Datta.
\newblock Mapping sub-second structure in mouse behavior.
\newblock \emph{Neuron}, 88\penalty0 (6):\penalty0 1121--1135, 2015.

\bibitem[Yingzhen \& Mandt(2018)Yingzhen and Mandt]{yingzhen2018disentangled}
Li~Yingzhen and Stephan Mandt.
\newblock Disentangled sequential autoencoder.
\newblock In \emph{International Conference on Machine Learning}, 2018.

\bibitem[Zhan et~al.(2020)Zhan, Tseng, Yue, Swaminathan, and
  Hausknecht]{zhan2020learning}
Eric Zhan, Albert Tseng, Yisong Yue, Adith Swaminathan, and Matthew Hausknecht.
\newblock Learning calibratable policies using programmatic style-consistency.
\newblock In \emph{International Conference on Machine Learning}, 2020.

\bibitem[Zhang et~al.(2006)Zhang, Ho, Zhang, and Lin]{zhang2006unsupervised}
Hui Zhang, Tu~Bao Ho, Yang Zhang, and M-S Lin.
\newblock Unsupervised feature extraction for time series clustering using
  orthogonal wavelet transform.
\newblock \emph{Informatica}, 30\penalty0 (3), 2006.

\bibitem[Zhang et~al.(2020)Zhang, Ti{\v{n}}o, Leonardis, and
  Tang]{zhang2020survey}
Yu~Zhang, Peter Ti{\v{n}}o, Ale{\v{s}} Leonardis, and Ke~Tang.
\newblock A survey on neural network interpretability.
\newblock \emph{arXiv preprint arXiv:2012.14261}, 2020.

\bibitem[Zhao et~al.(2017)Zhao, Song, and Ermon]{zhao2017learning}
Shengjia Zhao, Jiaming Song, and Stefano Ermon.
\newblock Learning hierarchical features from generative models.
\newblock In \emph{International Conference on Machine Learning}, 2017.

\end{thebibliography}
\bibliographystyle{tmlr}

\appendix
\appendix

\onecolumn

\section{Additional Results}

Table \ref{tab:metrics_stdev} contains the standard deviations of the results in Table \ref{tab:cluster_metrics} of the main paper.

Table \ref{tab:metrics_elbo} contains the median ELBO of our baselines and our neurosymbolic encoders. We find that our symbolic encoders are comparable with our baselines. This is expected: since we are imposing additional constraints on the encoder (a program with a bounded depth), we would not expect the variational approximation to be better than an encoder without these constraints (fully-neural encoder). In general, obtaining better or more semantically-meaningful cluster assignments can come at the cost of a smaller ELBO. For example, we find that introducing a clustering loss to the TVAE can result in better metrics, but in lower ELBO as well.

\begin{table*}[h]
\begin{center}
 \begin{tabular}{c|c c c | c c c} 
         \toprule[0.2em]
         \multirow{2}{*}{Model} & 
         \multicolumn{3}{c|}{\textbf{CalMS21}} &
         \multicolumn{3}{c}{\textbf{Basketball}}\\ 
        & Purity & NMI & RI  & Purity & NMI & RI  \\  \toprule[0.2em]
        TVAE  & .002 & .011 & .001 & .049 & .012 & .008 \\
        TVAE+KMeans loss & .001 & .002 & .001 & .006 & .001 & .001 \\ 
        JointVAE &  .000 & .003 & .022 & .037 & .020 & .004 \\
        VQ-TVAE &  .005 & .004 & .016 & .042 & .022 & .014 \\
        Beta-TVAE & .001 & .001 & .001 & .124 & .140 & .088 \\
        \hline
        Ours (1 program)  & .026 & .056 & .035 & .039 & .014 & .001 \\
        Ours (2 programs)  & .017 & .051 & .019 & .053 & .020 & .018 \\ 
        Ours (3 programs)  & .088 & .075 & .030 & .007 & .002 & .002 \\
        \hline
 \end{tabular}
\end{center}
\caption{Standard deviation of purity, NMI, and RI on CalMS21 and Basketball compared to human-annotated labels (3 runs). Random assignment metrics have standard deviation close to 0.  }
\label{tab:metrics_stdev}
\end{table*}

\begin{table*}[h]
\begin{center}
 \begin{tabular}{c|c | c} 
         \toprule[0.2em]
         Model & 
         \textbf{CalMS21} &
         \textbf{Basketball}\\ \toprule[0.2em]
        TVAE  & 1120 & 895 \\
        TVAE+KMeans loss & 1079 & 893 \\ 
        JointVAE &  1090 & 902\\
        VQ-TVAE &  971 & 911 \\
        Beta-TVAE &  1110 & 898\\
        \hline
        Ours (1 program)  &  1075 & 894\\
        Ours (2 programs)  & 1073 & 893\\ 
        Ours (3 programs)  & 1079 & 899 \\
        \hline
 \end{tabular}
\end{center}
\caption{Median ELBO of CalMS21 and Basketball across 3 runs. }
\label{tab:metrics_elbo}
\end{table*}

\begin{table}[]
    \centering
    \begin{tabular}{|c|c|c|c|c|c|c|c|}
        \hline
        & n. epochs & s. epochs & frontier size & penalty & max depth & lr & batch size \\
        \hline
        Synthetic & 10 & 10 & 30 & 0.01 & 2 & 0.0002 & 32 \\
        \hline
        CalMS21 & 6 & 10 & 8 & 0.01 & 5 & 0.001 & 256 \\
        \hline
        Basketball & 8 & 8 & 30 & 0.01 & 3 & 0.002 & 128  \\
        \hline
    \end{tabular}
    \caption{Hyperparameters for program learning. n. epochs and s. epochs represent the number of neural and symbolic epochs respectively, where the neural epoch is for the neural heuristic. lr is the learning rate.}
    \label{tab:hyperparams_prog}
\end{table}

\begin{table}[]
    \centering
    \begin{tabular}{|c|c|c|c|c|c|c|c|c|}
        \hline
        & epochs & z dim & h dim & RNN dim & adv. dim & disc. cap. & cont. cap. & lr  \\
        \hline
        Synthetic & 50 & 4 & 16 & 16 & 8 & 0.6 & - & 0.0002 \\
        \hline
        CalMS21 & 30 & 8 & 256 & 256 & 8 & 0.69 & 10 & 0.0001  \\
        \hline
        Basketball & 20 & 8 & 128 & 128 & 8 & 0.6 & 4 & 0.02 \\
        \hline
    \end{tabular}
    \caption{Hyperparameters for VAE training. The batch size is the same as the ones for program learning in Table~\ref{tab:hyperparams_prog}.}
    \label{tab:hyperparams_vae}
\end{table}

\begin{table}[t]
    \centering
    \begin{tabularx}{\linewidth}{ |c|Y|Y|Y|Y|Y|Y|Y| }
        \cline{2-8}
        \multicolumn{1}{c|}{} 
        & \multicolumn{3}{c|}{JointVAE}
        & \multicolumn{1}{c|}{VQ-TVAE} 
        & \multicolumn{3}{c|}{Beta-TVAE} \\
        \cline{2-8}
        \multicolumn{1}{c|}{} & weight & disc. cap & cont. cap & \# embeddings. & weight & cap & cap iter \\
        \hline
        CalMS21 & 100 & 0.69 & 10 & 4 & 100 & 20 & 10k \\
        \hline
        Basketball & 10 & 0.6 & 4 & 2 & 10 & 5 & 20k\\
        \hline
    \end{tabularx}
    \caption{Hyperparameters for baseline models. On CalMS21, the z dim for all baselines are 32 and trained for 200 epochs. }
    \label{tab:baseline_hyperparams}
\end{table}

\section{Implementation Details}

The hyperparameters for our approach are in Tables~\ref{tab:hyperparams_prog}, ~\ref{tab:hyperparams_vae} and the hyperparameters for baselines are in Table~\ref{tab:baseline_hyperparams}. We used the Adam~\cite{kingma2014adam} optimizer for all training runs. Specifically, Table~\ref{tab:hyperparams_prog} contains hyperparameters for program learning. Our use of the hyperparameters during the program learning process are the same as those from NEAR~\cite{shah2020learning}.
Table~\ref{tab:hyperparams_vae} contains the hyperparameters for training the VAE component of our model, including the hyperparameters we used for capacity.

\subsection{Baseline Details}

\textbf{TVAE.} We use a variation of the VAE where the inputs are trajectory data, called a TVAE \cite{co2018self, zhan2020learning, sun2020task}. Here, the neural encoder $\Qneural$ and decoder $p_{\theta}$ are instantiated with recurrent neural networks (RNN), where $\bfz \sim \Qneural(\cdot | \bfx)$. In this domain, $\bfx$ is a trajectory of length $T$: $\bfx = \{ x_1, \dots, x_T \}$. The TVAE objective is:
\begin{equation}
\begin{aligned}
\mathcal{L}^{\text{tvae}} = \mathbb{E}_{q_{\phi}} \bigg[ \sum_{t=1}^T -\log(p_{\theta}(x_{t} | x_{<t}, \mathbf{z})) \bigg] + D_{KL}(q_{\phi}(\mathbf{z} | \mathbf{x} ) || p(\mathbf{z})).
\end{aligned}
\label{eq:tvae_loss}
\end{equation}
All other baselines are variations of the TVAE, based on variations of VAE studied in recent works.

\textbf{TVAE + KMeans loss.} A few works~\cite{ma2019learning,luxem2020identifying} have studied adding a loss to the VAE framework to encourage clustering in the latent space, called the K-means loss. Given a data matrix $\bfz \in \mathbb{R}^{d \times N}$, the K-means objective is: 
\begin{equation}
\begin{aligned}
\mathcal{L}^{\text{k-means}} = \Trace(\bfz^T \bfz) - \Trace(\bfA^T \bfz^T \bfz \bfA),
\end{aligned}
\label{eq:k-means}
\end{equation}
where $\bfA \in \mathbb{R}^{N \times k}$ is called the cluster indicator matrix. We optimize this loss using the implementation in~\cite{luxem2020identifying}, where $\bfA$ is updated by computing the $k$-first singular values of $\sqrt{\bfz^t \bfz}$. The K-means loss is trained jointly with the TVAE loss (Eq~\ref{eq:tvae_loss}) as one of our baselines.

\textbf{JointVAE.} JointVAE~\cite{dupont2018learning} is a variation of VAE that jointly optimizes discrete (\bfc) and continuous (\bfz) latent variables. The JointVAE objective encourages the KL divergence terms to match capacities $C_z$ and $C_c$ that gradually increases during training. The objective is:
\begin{equation}
\begin{aligned}
\mathcal{L}^{\text{jointvae}} = \mathbb{E}_{q_{\phi}} [ \log p_{\theta}(\bfx | \bfz, \bfc) ] - \gamma | D_{KL}(q_{\phi}(\mathbf{z} | \mathbf{x} ) || p(\mathbf{z})) - C_z| - \gamma | D_{KL}(q_{\phi}(\mathbf{c} | \mathbf{x} ) || p(\mathbf{c})) - C_c| ,
\end{aligned}
\end{equation}
where $\gamma$ is a constant. Since the capacities of the discrete and continuous variables are controlled separately, the model is forced to encode information using both channels. Here, we use the trajectory formulation of JointVAE, where: 
\begin{equation}
\log p_{\theta}(\bfx | \bfz, \bfc) = \sum_{t=1}^{T} \log p_{\theta}(x_t | x_{<t}, \bfz, \bfc).
\end{equation}

\textbf{VQ-TVAE.} VQ-VAE~\cite{oord2017neural} combines vector quantization with VAEs. These models produce discrete latent encodings that are used to index an embedding table (or codebook). $\bfz_e$ , the continuous output of the encoder, is mapped to a discrete encoding based on its nearest neighbor in the codebook, then the indexed encoding $\bfz_q$ is used as input to the decoder. During training, the model learns the codebook, as well as the assignments. The objective is:
\begin{equation}
\begin{aligned}
\mathcal{L}^{\text{vqvae}} = \log p_{\theta}(\bfx | \bfz_q) + ||\text{sg}[\bfz_e] - e||_2^2 + \beta ||\bfz_e - \text{sg}[e]||_2^2 , 
\end{aligned}
\end{equation}
where $e$ are embeddings from the codebook, and sg represents the stopgradient operator.

\textbf{Beta-TVAE.} Beta-VAEs~\cite{higgins2016beta,burgess2018understanding} have been shown to learn disentangled representations from the image domain. As originally proposed, an adjustable hyperparameter $\beta$ is used to weigh the KL term in the VAE objective. We use the version of beta-vae training objective with gradually increasing capacity $C$ proposed in~\cite{burgess2018understanding}. This object is:
\begin{equation}
\begin{aligned}
\mathcal{L}^{\text{betavae}} = \mathbb{E}_{q_{\phi}} [ \log p_{\theta}(\bfx | \bfz) ] - \gamma | D_{KL}(q_{\phi}(\mathbf{z} | \mathbf{x} ) || p(\mathbf{z})) - C|,
\end{aligned}
\end{equation}
where $\gamma$ is a constant. Here, we apply the beta-VAE objective to trajectory data using the factorization shown in Eq~\ref{eq:cond_prob}.

\subsection{Metrics Definition}
We evaluate our programs quantitatively using standard cluster metrics relative to human-defined labels. The metrics we use are Purity~\citep{schutze2008introduction}, Normalized Mutual Information (NMI) \citep{zhang2006unsupervised}, and Rand Index (RI) \citep{rand1971objective}. These metrics have also been used by other works for evaluating clustering~\citep{ma2019learning,luxem2020identifying}. The definition of purity is:
\begin{equation}
\begin{aligned}
Purity = \frac{1}{n}\sum_{u \in U}\max_{v \in V} | u \cap v|
\end{aligned}
\end{equation}
where $U$ is the set of human-defined labels, $V$ is the set of cluster assignments from the algorithm, and $n$ is the total number of trajectories.

The NMI is defined as:
\begin{equation}
\begin{aligned}
NMI = \frac{\sum_{u \in U}\sum_{v \in V} |u \cap v| \log \big( \frac{n |u \cap v|}{|u| |v|} \big)}{\sqrt{\sum_{u \in U} |u| \log \frac{|u|}{n}\sum_{v \in U} |v| \log \frac{|v|}{n}}}
\end{aligned}
\end{equation}

RI is defined as:
\begin{equation}
\begin{aligned}
RI = \frac{TP + TN}{n(n-1)/2}
\end{aligned}
\end{equation}
where $TP$ are the number of trajectory pairs correctly placed into the same cluster, $TN$ are the number of trajectory pairs correctly placed into different clusters, and $n$ is the total number of trajectories. 
For all metrics, a value closer to 1 indicates clusters that more closely match the human-defined labels.

\section{Dataset and DSL Details}

\textbf{Synthetic.} 
We generate trajectories with the following steps:

\begin{enumerate}
    \item Sample initial position $x_1 \sim \calN([10, 10], [1,1])$.
    \item Sample velocity from $v = [v_x, v_y] \sim \calN([0, 0], [1,1])$ such that $0.05 < \norm{v}_2 < 0.4$.
    \item Sample force in $x$-direction $c_x \sim \text{Bernoulli}(0.5)$ and update $v_x' = v_x + 0.4 \cdot (2c_x - 1)$.
    \item Sample force in $y$-direction $c_y \sim \text{Bernoulli}(0.5)$ and update $v_y' = v_y + 0.4 \cdot (2c_y - 1)$.
    \item Generate trajectory with $x_{t+1} = x_t + v' + 0.2 \cdot \epsilon_t$, where $\epsilon_t \sim \calN(0,1)$. 
\end{enumerate}

$v'$ is fixed for an entire trajectory. $(c_x, c_y)$ defines a label for each trajectory (one of 4). The ground-truth decoder is linear with respect to $x, v, c_x, c_y$. The DSL for the synthetic dataset includes library functions that threshold the final $x$ and $y$ positions, used to demonstrate that the ground-truth can be learned and the information can be extracted from the neural latent space (Figure \ref{fig:synthetic}). Experiments were run locally with an Intel 3.6-GHz i7-7700 CPU with 4 cores and an NVIDIA GTX 1080 Ti GPU with 3584 CUDA cores.

\textbf{CalMS21.} The CalMS21 dataset~\cite{sun2021multi} consists of trajectory data from a mouse tracker~\cite{segalin2020mouse}, where each mouse is tracked by seven body keypoints from an overhead camera. The two mice are engaging in social interaction, where an intruder mouse is introduced to the cage of the resident mouse. The dataset contains an unlabelled split which we use for training and validation, and we use the test split of Task 1 in CalMS21 for testing. Each frame of the test split is annotated by a domain expert with one of four labels: attack, mount, investigation, other. We use these annotated behavior labels for comparison with clusters produced by our algorithm. This dataset is available under the CC-BY-NC-SA license.

The feature selects in the CalMS21 DSL are based on behavior attributes computed on trajectory data from domain experts in this area~\cite{segalin2020mouse}. In particular, we asked three domain experts to independently select features from~\cite{segalin2020mouse} to be part of the DSL. The time it takes domain experts to do this step is on the timescale of minutes. A full list of all features use in the DSLs are as follows:
\begin{itemize}
    \item Features in DSL 1: head body angle (resident and intruder), social angle (resident and intruder), speed (resident and intruder), distance between nose of resident and tail of intruder, distance between nose of resident and nose of intruder.
    \item Features in DSL 2: distance between head of mice, distance between body of mice, distance between head of resident to body of intruder, resident acceleration, resident nose speed, resident axis ratio of fitted ellipse, intersection over union of mice bounding boxes, resident social angle, distance between nose of resident and tail of intruder, distance between nose of resident and nose of intruder.
    \item Features in DSL 3: head body angle (resident and intruder), area of ellipse fitted to body keypoints (resident and intruder), acceleration (resident and intruder), distance between nose of resident and tail of intruder, distance between nose of resident and nose of intruder.
\end{itemize}
Note that unless otherwise stated, the CalMS21 experiments uses the features from DSL 1.

The experiments are ran on Amazon EC2 with an Intel 2.3 GHz Xeon CPU with 4 cores equipped with a NVIDIA Tesla M60 GPUs with 2048 CUDA cores. 

\textbf{Basketball.} 
The basketball dataset was also used in \cite{shah2020learning, zhan2020learning} and tracks the $xy$-positions of players from real NBA games. The positions are centered on the left half-court. Both (5) offensive and (5) defensive players are tracked, as well as the ball (excluded in our experiments). 

The DSL for basketball contains library functions that compute the speed, acceleration, final positions, and distance-to-basket of players and take the maximum, minimum, or average over the players. We did not consult a domain expert for this DSL, but these functions were used as labeling functions in \cite{zhan2020learning}. Basketball experiments were run locally with an Intel 3.6-GHz i7-7700 CPU with 4 cores and an NVIDIA GTX 1080 Ti GPU with 3584 CUDA cores.

% \section{Appendix}

\end{document}